\pdfoutput=1

\documentclass[11pt]{article}
\usepackage[dvipsnames]{xcolor}
\usepackage[final]{acl}
\usepackage{booktabs}
\usepackage{array}
\usepackage{times}
\usepackage{latexsym}
\usepackage[normalem]{ulem}
\useunder{\uline}{\ul}{}
\usepackage{graphicx}
\usepackage{booktabs}
\usepackage{pifont}
\usepackage{multirow}
\usepackage{subcaption}
\usepackage{float}

\usepackage{amsmath}
\usepackage{amsfonts}
\usepackage{amssymb}
\usepackage{mathtools}

\usepackage[T1]{fontenc}
\usepackage{hyperref}
\usepackage[utf8]{inputenc}

\usepackage{microtype}

\usepackage{inconsolata}

\usepackage{graphicx}

\usepackage{mathtools}
\title{\texttt{VisDoM}: Multi-Document QA with Visually Rich Elements\\ Using Multimodal Retrieval-Augmented Generation}

\author{Manan Suri $^ {\includegraphics[width=0.35cm]{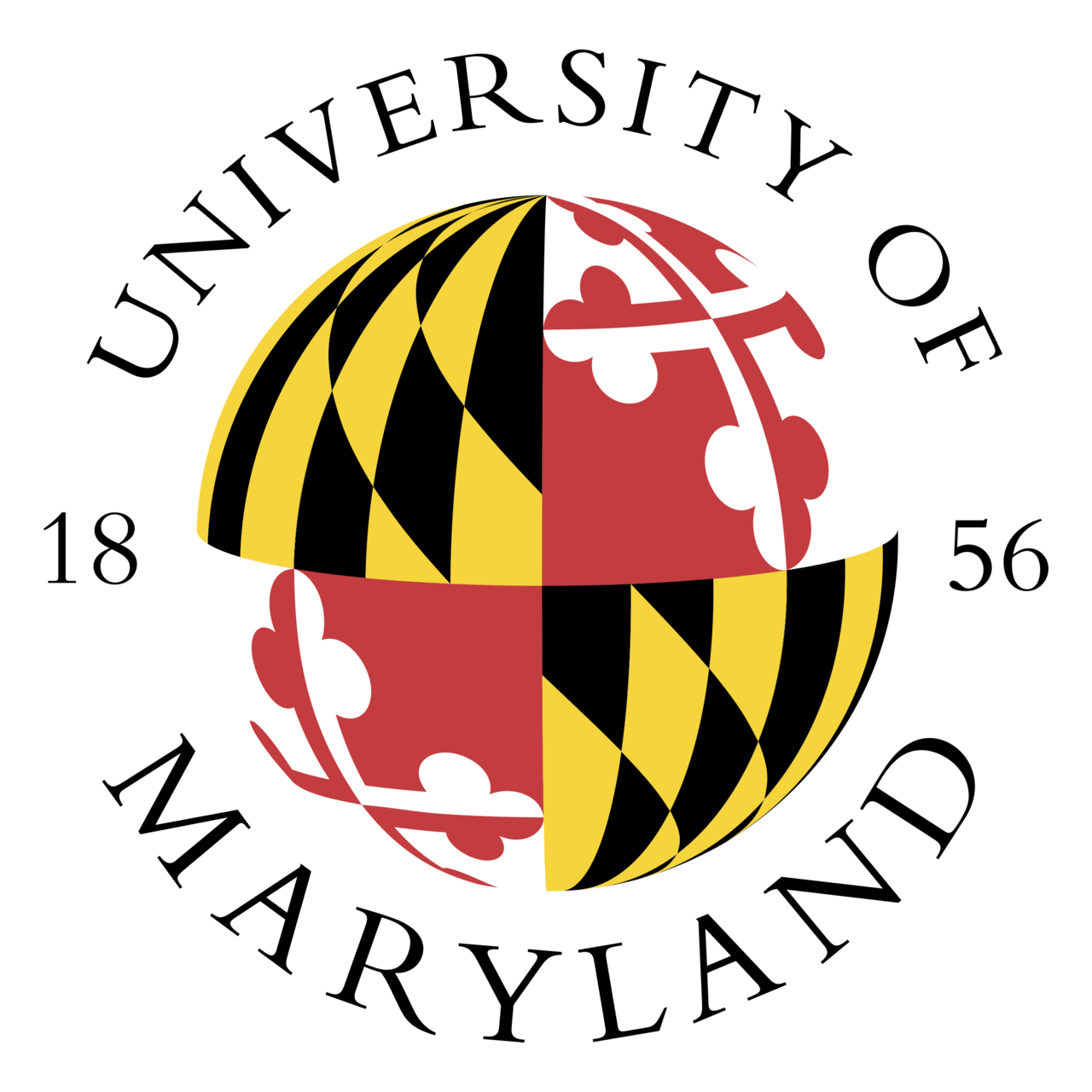}}$, \text{Puneet Mathur} $^ {{\includegraphics[width=0.2cm]{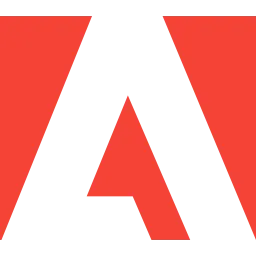}}}$ \thanks{Primary Research Mentor}, \text{Franck Dernoncourt}$ ^ {{\includegraphics[width=0.2cm]{figures/722666.png}} }$,\vspace{-5pt} \\ \textbf{\text{Kanika Gowswami}} $^ {\includegraphics[width=0.3cm]{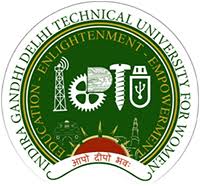}}$, \hspace{1.5pt} \textbf{\text{Ryan A. Rossi}} $^ {{\includegraphics[width=0.2cm]{figures/722666.png}} }$,\hspace{1.5pt} \textbf{ \text{Dinesh Manocha}} $^ {\includegraphics[width=0.35cm]{figures/university-of-maryland-logo-1.png}}$\\ $^{\includegraphics[width=0.35cm]{figures/university-of-maryland-logo-1.png}}$ $\text{University of Maryland, College Park}$ \hspace{10pt} $^{\includegraphics[width=0.2cm]{figures/722666.png}} \text{Adobe Research} $\hspace{10pt}  $^{\includegraphics[width=0.3cm]{figures/log.jpg}} $\text{IGDTUW} \\ \texttt{manans@umd.edu}, \texttt{puneetm@adobe.com}  }

\begin{document}
\maketitle
\begin{abstract}
Understanding information from a collection of multiple documents, particularly those with visually rich elements, is important for document-grounded question answering. This paper introduces \texttt{VisDoMBench}, the first comprehensive benchmark designed to evaluate QA systems in multi-document settings with rich multimodal content, including tables, charts, and presentation slides. We propose \texttt{VisDoMRAG}, a novel multimodal Retrieval Augmented Generation (RAG) approach that simultaneously utilizes visual and textual RAG, thereby combining robust visual retrieval capabilities with sophisticated linguistic reasoning. \texttt{VisDoMRAG} employs a multi-step reasoning process encompassing evidence curation and chain-of-thought reasoning for concurrent textual and visual RAG pipelines. A key novelty of VisDoMRAG is its consistency-constrained modality fusion mechanism, which aligns the reasoning processes across modalities at inference time to produce a coherent final answer. This leads to enhanced accuracy in scenarios where critical information is distributed across modalities and improved answer verifiability through implicit context attribution. Through extensive experiments involving open-source and proprietary large language models, we benchmark state-of-the-art document QA methods on \texttt{VisDoMBench}. Extensive results show that \texttt{VisDoMRAG} outperforms unimodal and long-context LLM baselines for end-to-end multimodal document QA by 12-20\%.
\end{abstract}


\section{Introduction}

In today's information-rich landscape, PDF documents play a crucial role in storing and disseminating information across various domains, including finance, legal, scientific research, and more. These documents often contain a rich blend of textual, visual, and tabular data, making them a unique challenge for information retrieval systems. Unlike structured formats like databases, PDFs are inherently unstructured, with diverse layouts combining paragraphs, images, charts, and tables. This complexity demands sophisticated multimodal processing techniques capable of interpreting both the textual and visual content. Effective handling of multimodal content from PDFs is essential for downstream tasks such as question-answering \cite{ding2022vdoc,mathew2021docvqa}, summarization \cite{pang-etal-2023-long}, and knowledge extraction \cite{pal-etal-2023-multitabqa}, where accurate and context-aware data extraction can significantly enhance decision-making processes. As a result, developing advanced methods that can fully leverage the multimodal nature of PDF documents has become a critical research challenge.

\begin{figure}
    \centering
    \includegraphics[width=\linewidth]{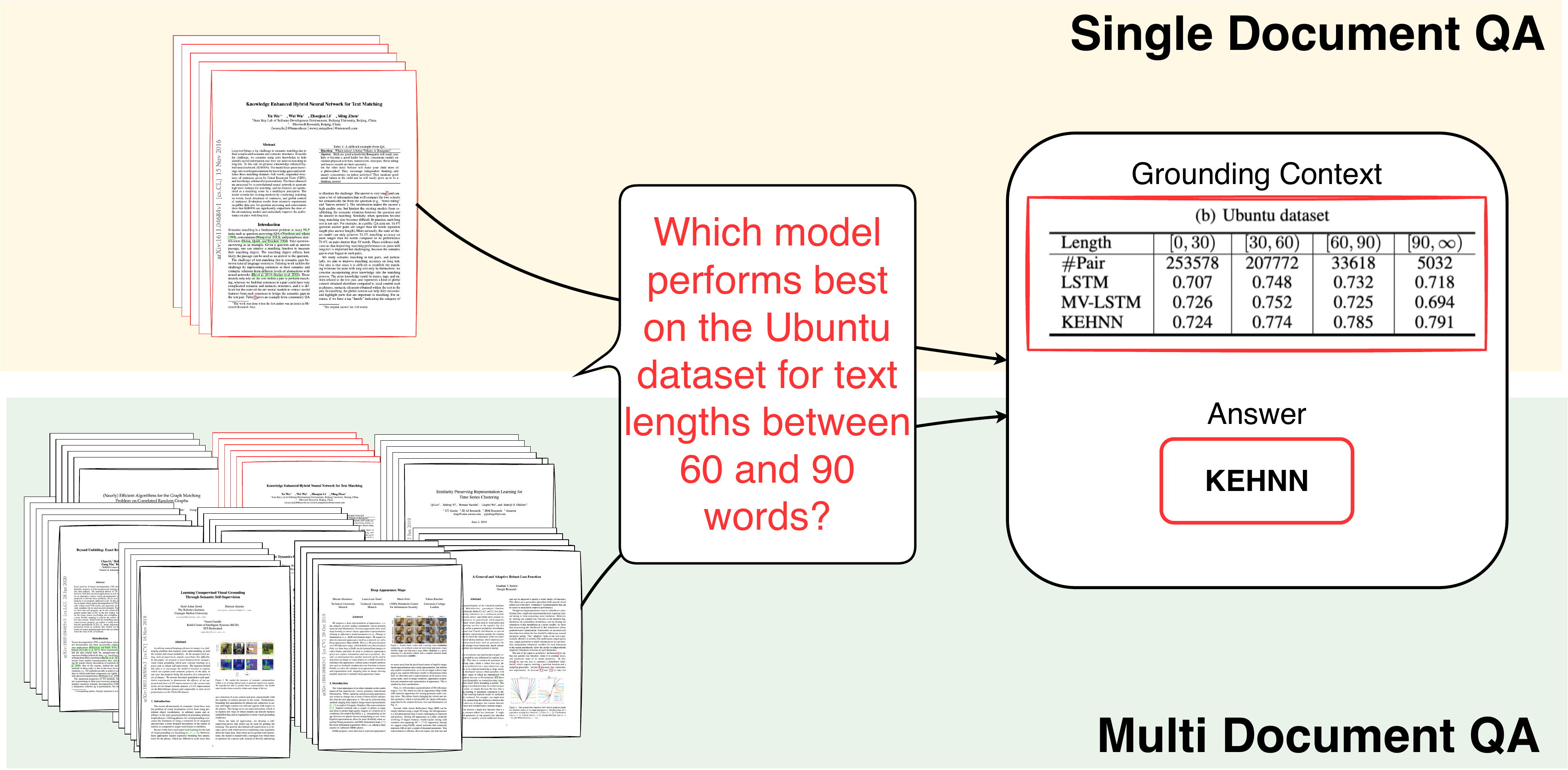}
    \caption{Multi-document QA systems require inferring relevant context from a large volume of unstructured data, inherently making it a more challenging task than single-document QA.}
    \label{fig:into}
\end{figure}

In real-world document QA systems, queries are often directed over a collection of source documents rather than a single source, requiring the system to identify the document that contains the relevant answer. This reflects common scenarios in domains such as finance, science, and policy analysis, where users interact with large, varied document sets to find specific information. In these cases, the challenge lies in effectively localizing context relevant to the query, from a large volume of information distributed across multiple documents (akin to finding a "needle in a haystack" \cite{wang2024multimodalniah}).

 Multi-document QA datasets are scarce, with existing multi-document benchmarks \cite{bai2023longbench, wang2024loong}, predominantly focused on textual information, often overlooking the diverse content forms found in real-world documents, such as tables, charts, and visual elements. Visually rich elements, such as tables, charts, and slides, provide structured data and visual summaries that are critical for answering certain types of questions. Tables often present dense, organized information that cannot be captured through plain text. At the same time, charts and slides can visually depict trends, relationships, or distributions that require interpretation beyond textual descriptions. The absence of datasets that include these modalities limits the ability of current QA models to address complex, multimodal questions. For instance, answering a financial or scientific question may require interpreting both numerical data in tables and trends in graphs alongside the surrounding text. 

In the context of visually rich content-based documents, existing RAG systems face a critical limitation due to their reliance on a singular modality (either text or vision) for retrieval. Text-based systems are proficient in linguistic reasoning but often overlook vital visual elements, such as tables and figures, that may contain key information. Conversely, multimodal RAG \cite{chen2022murag} systems that leverage vision-based retrieval can effectively extract visual data but are often constrained in end-to-end performance by the LLM's visual reasoning abilities, as text often performs better than visual input when given the same context \cite{deng2024tables}, which can be attributed to language bias in visual LLMs \cite{niu2021counterfactual, wang2024mdpo}, and visual hallucination \cite{ghoshvisual}.

\textbf{Main Results:} We introduce \texttt{VisDoMBench}, the first multi-document, multi-modal QA dataset specifically designed to address rich visual content, including tables, charts, and slides. \texttt{VisDoMBench} encompasses a diverse range of complex content and question types, along with annotated evidence, allowing for a comprehensive evaluation of multimodal QA systems. In this work, we benchmark the performance of various visual and textual retrieval methods on \texttt{VisDoMBench}, providing insights into their effectiveness in handling visually rich, multi-document queries. 

Further, we propose \texttt{VisDoMRAG}, a novel multimodal RAG approach that effectively performs modality fusion over textual and visual RAG pipelines, benefiting from the inherent strengths of both these approaches, unlike contemporary approaches, which perform only-text or only-vision-based retrieval. \texttt{VisDoMRAG} employs parallel RAG pipelines for text and visual elements, each with a multi-step reasoning process involving evidence curation, chain-of-thought reasoning, and answer generation. The system then integrates the outputs from both pipelines using modality fusion, which imposes a consistency constraint on the reasoning chains, ensuring inference-time alignment across the modalities' reasoning processes to produce the final answer. \texttt{VisDoMRAG} offers several significant advantages over traditional unimodal or simpler multimodal systems. Firstly, it ensures comprehensive information utilization by fully leveraging both textual and visual cues, leading to more accurate and complete answers, particularly in scenarios where critical information is distributed across different modalities. Moreover, the evidence curation step provides an additional advantage of answer verifiability, since context attribution is built into our approach. We conduct experiments utilizing various open-source and closed-source LLMs, comparing multiple strategies such as long-context processing, textual RAG, and visual RAG, with our proposed system.  We find that our \texttt{VisDoMRAG} improves end-to-end QA performance on our benchmarks, with performance gains in the range of 12\%-20\%. Our \textbf{main contributions} are:
\begin{itemize}
    \item \textbf{\texttt{VisDoMBench} \footnote{\url{https://github.com/MananSuri27/VisDoM/}}, a novel multi-document, multimodal QA benchmark} designed to address QA tasks across visually rich document content such as tables, charts, and slides, allowing for a comprehensive evaluation of multimodal document QA systems.
    
    \item \textbf{\texttt{VisDoMRAG}}, a \textbf{novel multimodal RAG} approach that effectively parallelly performs textual and visual RAG via Evidence Curation and Chain-of-Thought reasoning. The output reasoning chains from both the modalities are aligned using consistency analysis and resultant answers are ensembled together via LLM-based modality fusion to enhance visually-rich document QA.
    
    \item \textbf{\texttt{VisDoMRAG} significantly outperforms strong baselines} such as long-context processing, textual RAG, and visual RAG on the \textbf{\texttt{VisDoMBench} corpus by 12-20\%} across various open and closed-source LLM settings.
\end{itemize}

\begin{table*}[ht]
    \centering
    \resizebox{1.6\columnwidth}{!}{
    \begin{tabular}{lccc}
    \toprule
    \textbf{Benchmark} & \textbf{Content Type} & \textbf{Multi Document} & \textbf{Domain} \\ 
    \midrule
    L-Eval \cite{an2023leval} & Text & \textcolor{red}{\ding{55}} & Multi-domain \\ 
    LongBench \cite{bai2023longbench} & Text & \textcolor{OliveGreen}{\ding{51}} & Wikipedia \\ 
    Marathon \cite{zhang2023marathon} & Text & \textcolor{red}{\ding{55}} & Multi-domain \\ 
    LooGLE \cite{li2023loogle}& Text & \textcolor{red}{\ding{55}} & Multi-domain \\ 
    MPDocVQA \cite{tito2023hierarchical}& Text, Tables, Charts & \textcolor{red}{\ding{55}} & Multi-domain \\
    $\infty$Bench  \cite{zhang2024infty}& Text & \textcolor{red}{\ding{55}} & Multi-domain \\ 
    Ruler \cite{hsieh2024ruler}& Text & \textcolor{red}{\ding{55}} & Wikipedia \\ 
    Loong \cite{wang2024loong}& Text & \textcolor{OliveGreen}{\ding{51}} & Multi-domain \\ 
    UDA \cite{hui2024uda}& Text, Tables & \textcolor{red}{\ding{55}} & Multi-domain \\ 
    NarrativeQA \cite{kovcisky2018narrativeqa}& Text & \textcolor{red}{\ding{55}} & Movies and Shows \\ 
    MMLONGBENCH-DOC \cite{ma2024mmlongbench}& Text, Tables, Charts, Slides & \textcolor{red}{\ding{55}} & Multi-domain \\ 
     
    \midrule
    \textbf{\texttt{VisDoMBench} (Ours)} & \textbf{Text, Tables, Charts, Slides} & \textbf{\textcolor{OliveGreen}{\ding{51}}} & \textbf{Multi-domain} \\ 
    \bottomrule
    \end{tabular}}
    \caption{Comparison of long context document QA benchmarks with \texttt{VisDoMBench}.}
    \label{tab:benchmarks}
\end{table*}

\section{Related Work}

\textbf{Retrieval Augmented Generation}
While Large Language Models (LLMs) have achieved significant advancements, they still encounter challenges in integrating external knowledge and adapting to new, unseen data. Retrieval Augmented Generation (RAG) addresses these gaps by incorporating external information, enhancing the precision and reliability of LLM responses \cite{lewis2020rag}. RAG is utilized across various downstream unimodal NLP tasks, including machine translation \cite{gu2018ragmt1,he2021ragmt2}, dialogue generation \cite{cai2018ragdialog}, abstractive summarization \cite{peng2019ragsummary}, and knowledge-intensive generation \cite{izacard2020ragknowledge,lewis2020rag}. In visual question answering (VQA), \cite{lin2022ragvqa} addresses open-domain challenges by using object detection, image captioning, and optical character recognition (OCR) to transform target images into textual data. Moving beyond text-only contexts, MuRAG retrieves both text and image data, incorporating images as visual tokens \cite{chen2022murag}. RAMM enhances performance by retrieving and encoding similar biomedical images and their captions through distinct networks \cite{yuan2023ramm}. 


\noindent{\textbf{Long Context Document Benchmarks}}
The comparison of long context document question-answer benchmarks (Table \ref{tab:benchmarks}), highlights the diversity in content types, multi-document capabilities, and domains. Existing benchmarks such as L-Eval \cite{an2023leval}, Marathon \cite{zhang2023marathon}, and LooGLE \cite{li2023loogle} primarily focus on text-based content from multi-domain sources but do not support multi-document inputs. LongBench \cite{bai2023longbench} and Loong \cite{wang2024loong} extend their evaluations to include multi-document settings, although they remain text-centric.

\noindent\textbf{Comparison with existing datasets}: Certain benchmarks like MPDocVQA \cite{tito2023hierarchical}, UDA \cite{hui2024uda}, and MMLONGBENCH-DOC \cite{ma2024mmlongbench} expand the content spectrum by incorporating tables, charts, and slides, but they are limited to single-document question answering. In contrast, \texttt{VisDoMBench} supports multi-document question answering across various content types, including text, tables, charts, and slides, offering a more comprehensive multi-domain evaluation framework.


\begin{table*}[h!]
    \centering
    \resizebox{2\columnwidth}{!}{
    \begin{tabular}{lccccccccccc}
        \hline
        \textbf{Dataset} & \textbf{Domain} & \textbf{Content Type} & \textbf{Queries} & \textbf{Docs} & \textbf{Avg. Question Length} & \textbf{Avg. Doc Length \small{(Pages)}} & \textbf{Avg. Docs per Query} & \textbf{Avg. Pages per Query} \\
        \hline
        \textbf{PaperTab} & Wikipedia & Tables, Text & 377 & 297 & 29.44 \small{$\pm 6.3$} & 10.55 \small{$\pm 6.3$} & 10.82 \small{$\pm 4.4$} & 113.10 \small{$\pm 50.4$} \\
        \textbf{FetaTab} & Scientific Papers & Tables & 350 & 300 & 12.96 \small{$\pm 4.1$} & 15.77 \small{$\pm 23.9$} & 7.77 \small{$\pm 3.1$} & 124.33 \small{$\pm 83.0$} \\
        \textbf{SciGraphQA} & Scientific Papers & Charts & 407 & 319 & 18.05 \small{$\pm 1.9$} & 22.75 \small{$\pm 29.1$} & 5.91 \small{$\pm 2.0$} & 129.71 \small{$\pm 81.7$} \\
        \textbf{SPIQA} & Scientific Papers & Tables, Charts & 586 & 117 & 16.06 \small{$\pm 6.6$} & 14.03 \small{$\pm 7.9$} & 9.51 \small{$\pm 3.5$} & 135.58 \small{$\pm 55.2$} \\
        \textbf{SlideVQA} & Presentation Decks & Slides & 551 & 244 & 22.39 \small{$\pm 7.8$} & 20.00 \small{$\pm 0.0$} & 6.99 \small{$\pm 2.0$} & 139.71 \small{$\pm 40.6$} \\
        \hline
        \textbf{\texttt{VisDoMBench}} & Combined & Tables, Charts, Slides, Text & 2271 & 1277 & 19.11 \small{$\pm 5.4$} & 16.43 \small{$\pm 14.5$} & 8.36 \small{$\pm 3.0$} & 128.69 \small{$\pm 62.7$} \\
        \hline
    \end{tabular}}
    \caption{Summary of data splits included in \texttt{VisDoMBench}.}
    \label{tab:summary1}
\end{table*}



\section{Problem Formulation}

Given a query $q$, we have a collection of $ M $ documents $ \mathcal{D} = \{d_{1}, d_{2}, \dots, d_{M}\} $, wherein each document $ d_{i} $ may consist of a set of $N_i$ pages represented by $ P^{i} = \{p^{i}_{1}, p^{i}_{2}, \dots, p^{i}_{N_i}\} $. We aim to generate text $\hat{a}$ for each query $q$ that accurately answers the user query. The answer generation relies on retrieving relevant evidence context from one or more documents. Each query $q$ may require information spread across different pages from one or more of the associated documents in $D$.

\noindent We aim to propose a framework that can accurately answer questions over a collection of multi-page documents where the system first retrieves relevant evidence at the level of individual pages, paragraphs or text chunks, followed by using the retrieved context to generate answer text.



\section{\texttt{VisDoMBench}}

Every data point in VisDoMBench can be expressed as triple $(q, D, \hat{a})$, where a question $q$ is posed to a set of documents $D$, with ground-truth answer $\hat{a}$. We re-purpose five existing document-QA datasets to form our benchmark. Table \ref{tab:summary1} summarises different data splits present in \texttt{VisDoMBench}, including summary statistics, QA type, and content type.

\subsection{\texttt{VisDoMBench}}

\noindent\textbf{Data Sourcing}: In the curation of document question-answering datasets, we adhered to the following criteria: (1) the inclusion of visually rich content, encompassing tables, charts, and presentation slides; (2) the utilization of publicly accessible source documents; and (3) the presence of grounded evidence. These parameters were established to ensure the datasets' relevance to multimodal information retrieval and their applicability to real-world question-answering tasks. Our corpus comprises test/eval sets sourced from several established datasets. We incorporated the PaperTab and FeTaTab splits from the UDA Benchmark\cite{hui2024uda}, which in turn sourced these datasets from QASPER\cite{qasper} and FeTaQA\cite{nan2022fetaqa}, respectively. For chart-based question-answering samples, we drew from SciGraphQA \cite{li2023scigraphqa}, which is multi-turn QA dataset on charts from scientific papers, and SPIQA\cite{pramanick2024spiqa}, a chart and table QA dataset system sourced from \cite{qasper}. Additionally, we included SlideVQA \cite{tanaka2023slidevqa}, a multi-image, multi-hop QA dataset centered on presentation slide decks.

\begin{figure*}[h]
    \centering
    \includegraphics[width=\linewidth]{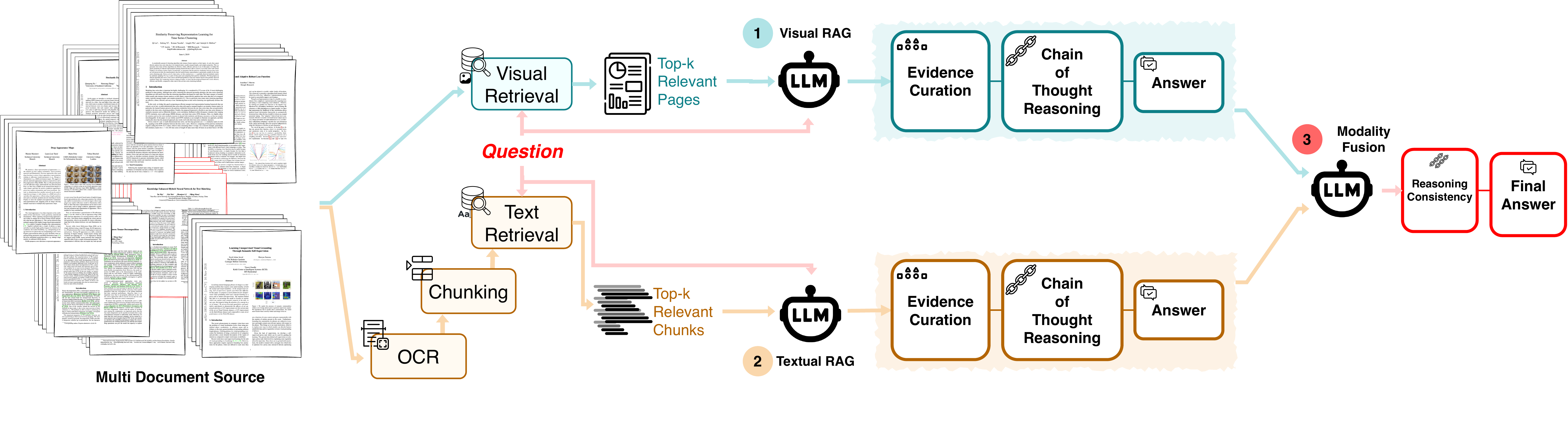}
    \caption{\texttt{VisDoMRAG}: Given a set of documents, VisDoMRAG parallelly performs evidence-driven \textcolor[HTML]{0E8088}{\ding{202}} Visual RAG and \textcolor[HTML]{B46504}{\ding{203}} Textual RAG, prompting the LLMs to answer a query based on the respective retrieved context via Evidence Curation and Chain-of-Thought reasoning. The reasoning chains, and answers from the text and visual pipeline are ensembled together via \textcolor[HTML]{FF0000}{\ding{204}} Modality Fusion, where the outputs of both the modalities are aligned using consistency analysis on their reasoning chain to arrive at the final answer.}
    \label{fig:main}
\end{figure*}

\noindent\textbf{Data Sampling}: Sourced QA pairs need to be sampled to retain high quality samples. To maintain the integrity and uniqueness of our benchmark, we meticulously removed overlapping samples between PaperTab and SPIQA and implemented rigorous de-duplication of QA pairs across all included datasets. Further, we also perform question-level de-duplication to ensure similar questions are not repeated across different document collections. This ensures that QA systems are not rewarded disproportionately for better handling particular question types. For \texttt{SciGraphQA}, we filter out trivial questions related to layout and document metadata. From the remaining questions, we randomly sample 500 questions from the top 50\%-ile of questions by length. The rationale for filtering on answer length filter is based on the heuristic that longer questions tend to be more specific, making them better suited for multi-document QA tasks, where specificity is crucial. For \texttt{SlideVQA}, we exclude single-hop questions, as they are generally non-specific and may have more than one correct answer from the document collection. We heuristically observe that multi-hop questions in this dataset are more likely to reference content from specific documents, thus making them a better fit for multi-document setups. SciGraphQA and SPIQA contain questions specific to charts or tables extracted from scientific papers. We use the ArXiv API\footnote{https://info.arxiv.org/help/api/index.html} to extract full document PDFs.

\noindent{\textbf{Document Augmentation}}: To simulate realistic multi-document settings, we augment each question across all data splits with varying number of distracting documents, ($|\mathcal{D}_i=M|$). We intend to keep the expected number of total pages per query between 50 to 200 to ensure that there is sufficient distracting content while maintaining the practical feasibility of contemporary long-context models. Hence, based on the average number of pages per document $P_{avg}$, we randomly sample the number of distracting documents $l$ to lie between the range $[\lfloor \frac{50}{P_{avg}} \rfloor, \lfloor \frac{200}{P_{avg}} \rfloor]$. Randomly sampling $l$ ensures that each benchmark instance contains a diverse degree of multi-document evidence, allowing for a more thorough evaluation of the QA model's retrieval and reasoning capabilities. 

\noindent{\textbf{Query Augmentation}}: To address the challenge of ambiguous questions in datasets such as \texttt{SciGraphQA}, and \texttt{PaperTab}, we implement a query augmentation procedure to create a one-to-one mapping between a given question and the document(s) that exclusively answer it. Given an original question and the document containing answer, we utilize GPT-4o to generate more specific variations of the question, ensuring that the generated question can only be answered by the corresponding document. To maintain consistency, we constrain the LLM such that the answer to the generated question must match the provided answer. Once the augmented queries are generated, a human annotator reviews them using a predefined rubric. The rubric guides the annotator to either select one of the five generated questions, retain the original question, or mark all questions (synthetic and actual) as ambiguous, in which case, the data point is discarded. The annotator is tasked with ensuring that the question is sufficiently specific by cross-referencing the localized evidence. Additionally, the annotator performs a simple search across the entire document collection to verify that the question cannot be ambiguously answered by any other document. Experimental validation of one-to-one mapping of query with respect to the source document is given in the Appendix.

\section{\texttt{VisDoMRAG}}
\label{sec:visdomrag}

\texttt{VisDoMRAG} (Fig \ref{fig:main}) is a multimodal RAG approach for visually rich document QA consisting of two steps: (i) parallel evidence-driven unimodal (vision and textual) RAG pipelines, and (ii) Modality Fusion, which imposes consistency constraints to combine unimodal reasoning chains and arrive at a final answer.

\subsection{Evidence-driven Parallel Unimodal RAG}

\paragraph{Textual Retrieval Pipeline}
The textual RAG pipeline commences with the extraction of text from the set of documents utilizing Optical Character Recognition (OCR), followed by the segmentation of the extracted text into smaller, indexable chunks. Metadata indicating the source document and page number is preserved to facilitate traceability. These chunks are then indexed using a text embedding model, enabling efficient retrieval. Relevant chunks are subsequently retrieved in relation to the specified query by a text retrieval model and provided as contextual input to the LLM along with the query to generate textual answer response.

\paragraph{Visual Retrieval Pipeline}
Simultaneously, the visual RAG pipeline is dedicated to the extraction and analysis of graphical elements, including images, charts, and diagrams. For a given set of PDFs, a visual embedding model generates an index at the page-level granularity for all documents. Relevant pages are then retrieved by a visual retrieval model based on the specified query, and these pages are supplied to multimodal LLMs as visual context. This approach ensures that the model has access to critical visual information, employing its multimodal capability to utilize visual cues from document layout and graphical structures such as charts, diagrams and infographics.

\noindent{\textbf{Prompting Strategy}} Both the textual and visual pipelines employ a sophisticated three-step prompting strategy. Given a set of context artifacts (page images or textual chunks), and a query, the LLM is prompted with the following steps:

\noindent\textbf{1. Evidence Curation}: As a first step, we prompt the LLM to extract relevant evidence from the retrieved context. The LLM must isolate key sections, such as paragraphs, tables, and figure details, that are most likely to address the query and verbalize them in a structured form. This curation is crucial in a multi-document setup, where non-uniform sources introduce irrelevant, distracting, or adversarial content. Accurately identifying relevant information enhances the model’s reasoning abilities by filtering out noise and helps mitigate LLM hallucinations.

\noindent\textbf{2. Chain of Thought Reasoning}: Extracting reasoning chains from multi-document artefacts can help contextualize curated evidence for final answer generation. We utilize Chain-of-Thought (CoT) \cite{wei2022chain} reasoning to link individual pieces of evidence that form a coherent step-by-step narrative, ensuring that the answer is not only accurate but also logically derived from the evidence, leading to more robust and reliable responses.

\noindent\textbf{3. Answer Generation}: By leveraging insights from curated, contextually relevant evidence and applying CoT reasoning processes, the answer generation step produces responses that are both precise and well-justified. Additionally, we use targeted prompts to guide the LLM about the appropriate format for answer generation as per the question type.

\begin{figure*}[h!]
    \centering
    \begin{subfigure}{0.24\textwidth}
        \centering
        \includegraphics[width=\linewidth]{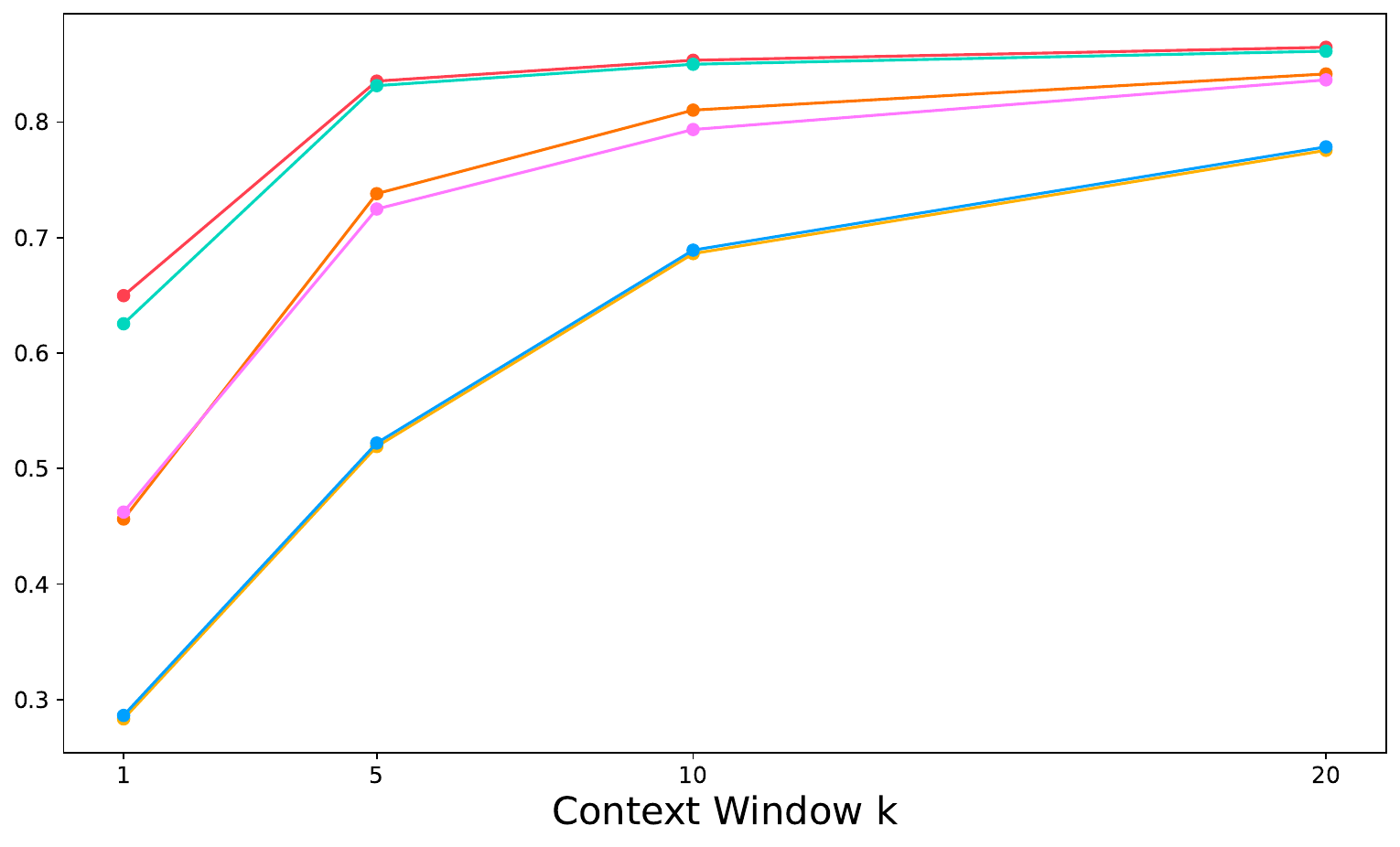}
        \caption{PaperTab}
    \end{subfigure}
    \begin{subfigure}{0.24\textwidth}
        \centering
        \includegraphics[width=\linewidth]{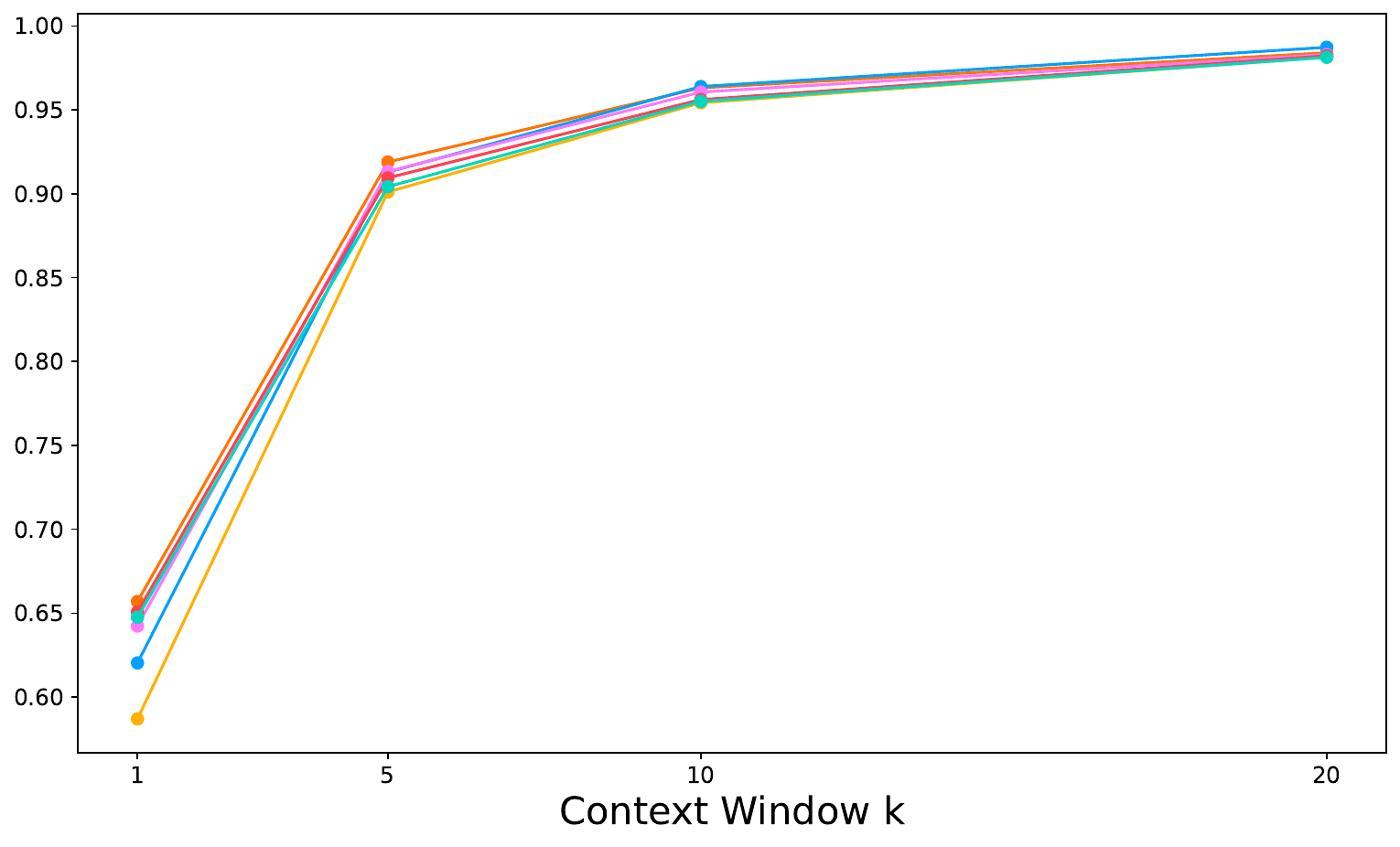}
        \caption{FetaTab}
    \end{subfigure}
    \begin{subfigure}{0.24\textwidth}
        \centering
        \includegraphics[width=\linewidth]{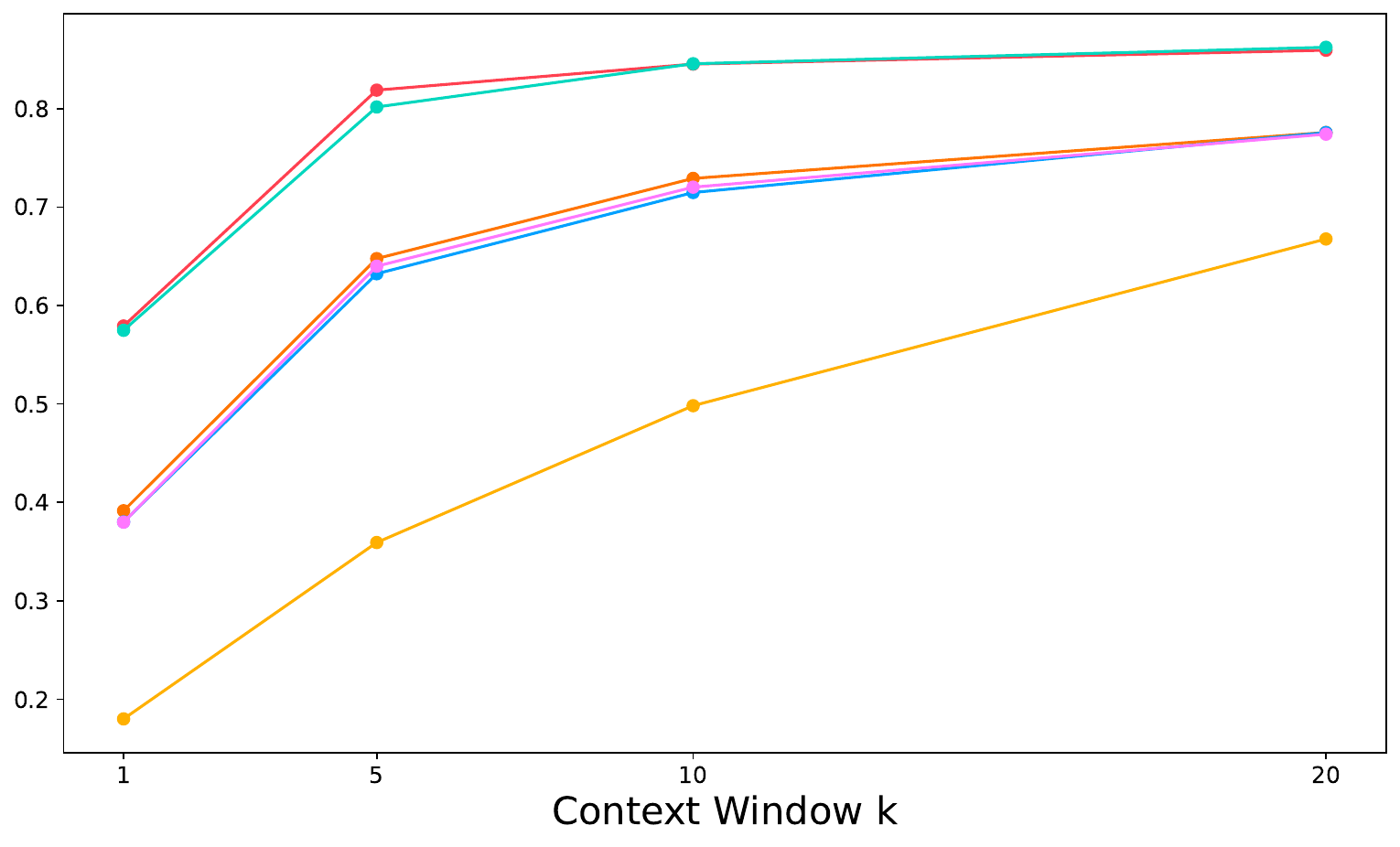}
        \caption{SciGraphQA}
    \end{subfigure}
    \begin{subfigure}{0.24\textwidth}
        \centering
        \includegraphics[width=\linewidth]{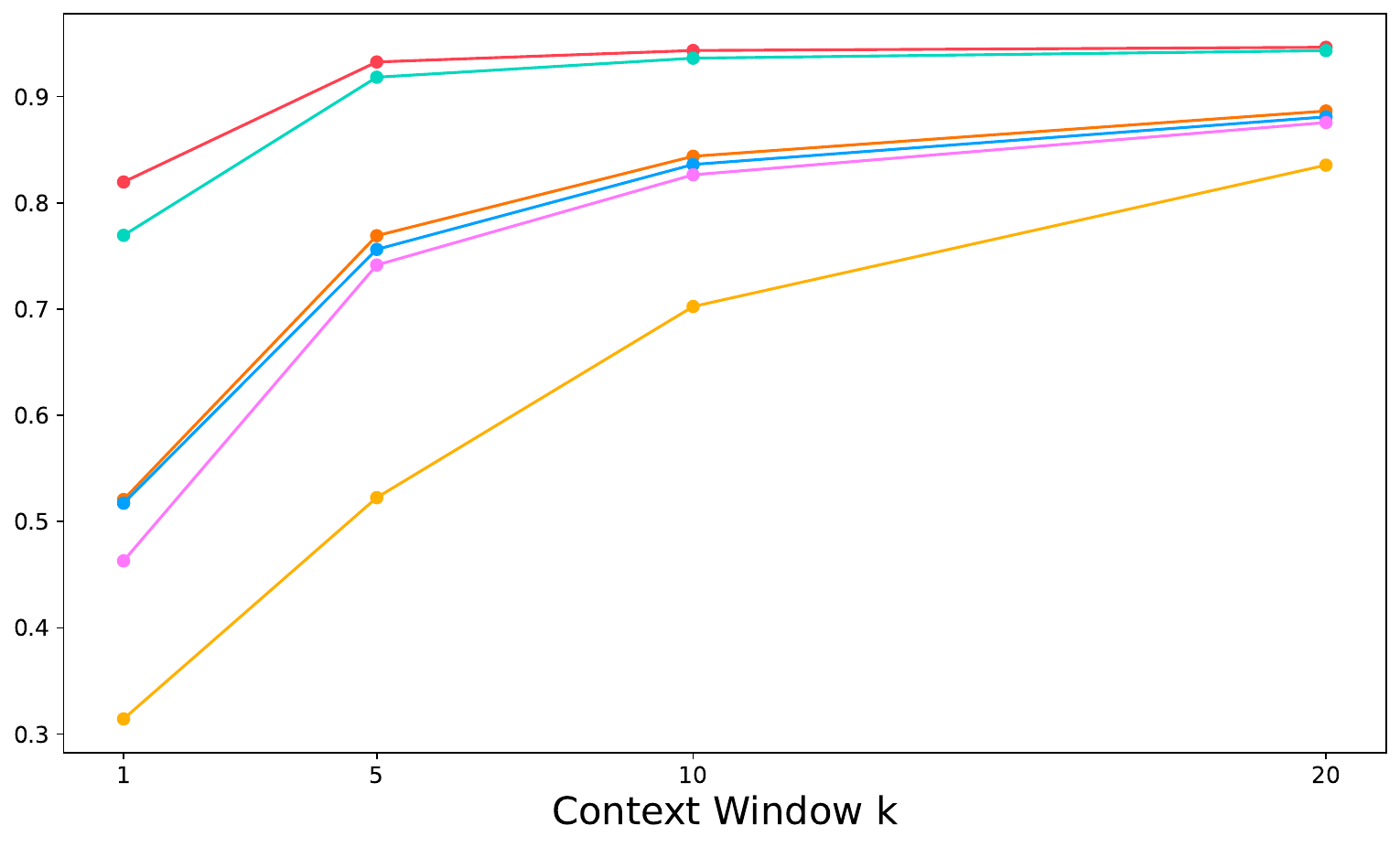}
        \caption{SPIQA}
    \end{subfigure}
        \caption{Comparison of retrieval performance across datasets, for benchmarked retrievers (\textcolor[HTML]{FFB000}{BM25}, \textcolor[HTML]{00A0FF}{MiniLM}, \textcolor[HTML]{FF77FF}{MPNet}, \textcolor[HTML]{FF7400}{BGE1.5}, \textcolor[HTML]{00D7BE}{ColPali}, \textcolor[HTML]{FF4050}{ColQwen}), at different context window lengths, varying $k \in [1,5,10,20]$.
    }
    \label{fig:four_subfigures}
\end{figure*}
\subsection{Modality Fusion}

The modality fusion stage is a key contribution in \texttt{VisDoMRAG} which differentiates it from simpler multimodal approaches. This stage takes as input the outputs from both the textual and visual pipelines, including the curated evidence, reasoning chains, and generated answers. The fusion process is orchestrated by prompting an LLM to evaluate the consistency between the reasoning chains produced by the textual and visual pipelines. This idea is inspired by self-consistency in CoT \cite{wang2023selfconsistencyimproveschainthought}, which leveraged multiple thought-chains and derives an answer based on the consistency of the individual chains' results. Consistency constraint prompting is crucial for identifying and resolving any discrepancies, contradictions and filling in reasoning gaps that may arise from the separate processing of different modalities. When inconsistencies are detected, the LLM is tasked with reconciling the differences, potentially by re-evaluating the evidence or adjusting the reasoning steps. This process ensures that the final answer integrates information from both modalities in a coherent and logically consistent manner.

\begin{table*}[]
\centering
\resizebox{1.6\columnwidth}{!}{%
\begin{tabular}{llcccccc}
\hline
\textbf{Baseline} & \textbf{LLM} & \multicolumn{1}{l}{\textbf{PaperTab}} & \multicolumn{1}{l}{\textbf{FetaTab}} & \multicolumn{1}{l}{\textbf{SciGraphQA}} & \multicolumn{1}{l}{\textbf{SPIQA}} & \multicolumn{1}{l}{\textbf{SlideVQA}} & \multicolumn{1}{l}{\textbf{Average}} \\ \hline
 & \textbf{Qwen2-VL} & 8.23 & 23.1 & 16.74 & 9.93 & 2.46 & 12.09 \\
 & \textbf{Gemini} & 27.62 & 62.02 & 22.1 & 38.82 & 13.47 & 32.81 \\
\multirow{-3}{*}{\textbf{Long Context}} & \textbf{GPT4o} & 28.37 & 60.03 & 24.12 & 36.3 & 15.06 & 32.78 \\ \hline
{\color[HTML]{B46504} } & \textbf{Qwen2-VL} & 25.33 & 57.56 & 26.75 & 39.77 & 8.82 & 31.65 \\
{\color[HTML]{B46504} } & \textbf{Gemini} & 33.6 & 63.86 & 26.48 & 42.33 & 10.3 & 35.31 \\
\multirow{-3}{*}{{\color[HTML]{B46504} \textbf{Text RAG}}} & \textbf{ChatGPT4o} & 37.34 & 60.82 & 29.74 & 42.8 & 15.97 & 37.33 \\ \hline
{\color[HTML]{0E8088} } & \textbf{Qwen2-VL} & 27.37 & 58.57 & 28.13 & 42.81 & 38.42 & 39.06 \\
{\color[HTML]{0E8088} } & \textbf{Gemini} & 29.23 & 52.82 & 23.56 & 41.43 & 51.96 & 39.80 \\
\multirow{-3}{*}{{\color[HTML]{0E8088} \textbf{Visual RAG}}} & \textbf{ChatGPT4o} & \underline{42.01} & \underline{61.89} & \underline{31.12} & \underline{43.28} & \underline{66.82} & \underline{49.02} \\ \hline
{\color[HTML]{CB0000} } & \textbf{Qwen2-VL} & 29.89 & 59.24 & 27.98 & 42.8 & 39.77 & 39.94 \\
{\color[HTML]{CB0000} } & \textbf{Gemini} & 39.66 & 60.89 & 25.82 & 41.03 & 52.74 & 44.03 \\
\multirow{-3}{*}{{\color[HTML]{CB0000} \textbf{VisDoMRAG}}} & \textbf{ChatGPT4o} & \textbf{44.11} & \textbf{63.28} & \textbf{31.36} & \textbf{44.09} & \textbf{67.22} & \textbf{50.01} \\ \hline
\end{tabular}%
}
\caption{Performance of our approach, \texttt{VisDoMRAG}, compared to baseline approaches on \texttt{VisDoMBench}. \texttt{VisDoMRAG} outperforms long-context LLM, visual and text-only RAG baselines.}
\label{tab:results}
\end{table*}

\section{Experiments}

In our experiments, we first evaluate different retrieval and indexing models on our benchmark, followed by end-to-end QA evaluation using the identified optimal retrieval models with different LLMs. The experiments, baselines and evaluation are discussed below:

\subsection{Retrieval }
\textbf{Baselines:} We use popular text based retrieval models: BM25 \cite{robertson1995bm25} a statistical baseline, and , MPNet \cite{mpnet},  MiniLM \cite{wang2020minilmdeepselfattentiondistillation}, and BGE-1.5 \cite{bge_embedding}, which represent SoTA dense retrieval baselines. Text extraction from PDF documents is performed using \texttt{PyTesseract}. The extracted text is then segmented into 3000-character chunks using the recursive-split method \cite{sarmah2023reducinghallucinationextractinginformation}, with a 10\% overlap to mitigate information loss.

For visual retrieval, we utilize recent advances late interaction based multi-vector retrieval models built on top of LLMs \cite{faysse2024colpali}, namely ColPali and ColQwen2, which have PaliGemma \cite{beyer2024paligemmaversatile3bvlm} and Qwen2 \cite{yang2024qwen2technicalreport} as their base LLMs. Readers are encouraged to refer to the appendix for further details of these models.

\noindent{\textbf{Evaluation:}} Evidence extraction is assessed using ANLCS between ground truth evidence and retrieved chunks/pages. Document identification evaluates the retrievers' ability to select the correct source document in a multi-document setup. We report the rate of instances where the ground truth document is the source of the majority  of the retrieved context.

\begin{table}[]
\resizebox{\columnwidth}{!}{%
\begin{tabular}{lcccccc}
\hline
\textbf{Retriever} &
  \multicolumn{1}{l}{\textbf{PaperTab}} &
  \multicolumn{1}{l}{\textbf{FetaTab}} &
  \multicolumn{1}{l}{\textbf{SciGraphQA}} &
  \multicolumn{1}{l}{\textbf{SPIQA}} &
  \multicolumn{1}{l}{\textbf{SlideVQA}} &
  \multicolumn{1}{l}{\textbf{Average}} \\ \hline
{\color[HTML]{FFB000} \textbf{BM25}}     & 65.51 & 84.00 & 72.73                & 88.23 & \textbf{98.55} & 81.80 \\
{\color[HTML]{00A0FF} \textbf{MiniLM}}   & 65.51 & 88.85 & 91.65                & 61.06 & 0.73  & 61.56 \\
{\color[HTML]{FF77FF} \textbf{MPNet}}    & 90.18 & 89.71 & 91.40                & 95.84 & 0.73  & 73.57 \\
{\color[HTML]{FF7400} \textbf{BGE1.5}}   & 96.81 & 94.00 & 90.91                & \textbf{98.43} & 81.85 & 92.40 \\
{\color[HTML]{00D7BE} \textbf{ColPali}}  & 


96.93&	\textbf{97.71}	&95.28	&93.17&	97.64	&96.15 \\
{\color[HTML]{FF4050} \textbf{ColQwen2}} & \textbf{97.61} & 96.86 & \textbf{95.58}                & 96.85 & 97.82 & \textbf{96.94} \\ \hline
\end{tabular}%
}
\caption{Comparison of performance in source document identification, at $k=5$.}
\label{tab:aux-res1}
\end{table}



\subsection{End-to-End QA}
We use the best text and visual retrieval models from the retrieval experiments for End-to-End QA evaluation. 

\noindent{\textbf{Baselines:}} We benchmark our method using LLMs capable of handling multi-image inputs and long context. To this extent, we include two off-the-shelf models Gemini-1.5-Flash \cite{reid2024gemini}, and ChatGPT-4o \cite{openai_gpt4o_2024}, as well as Qwen2-VL-7B-Instruct \cite{yang2024qwen2technicalreport}, an open-source LLM with visual and long context capabilities. We evaluate these LLMs in four approaches: 1. Long Context: where text content of all documents queries for a sample is passed as context, and 2. TextualRAG, 3. VisualRAG, and, 4.VisDoMRAG as described in Section \ref{sec:visdomrag}.

\noindent{\textbf{Evaluation:}} For PaperTab, we borrow the modified implementation of Word Overlap F1 from \cite{hui2024uda}, which takes into account different answer types (binary, short text). For all other datasets, we report the Word Overlap F1, which serves as a flexible metric to evaluate different answer types.

\section{Results}


\subsection{Retrieval Evaluation on \texttt{VisDoMBench}}

\noindent Fig. \ref{fig:four_subfigures} presents the performance of various retrieval models in extracting evidence from documents, evaluated using the Averaged Normalized Longest Common Subsequence (ANLCS) between retrieved evidence and ground truth evidence, for different context window lengths ($k = [1,5,10,20]$). Based on a threshold of ANLCS = 0.7,  we use a context window of $k=5$, $k=7$ for Visual RAG and Textual RAG, respectively, with ColQwen2 and BGE-1.5 as the visual and textual retrievers. ColQwen2 outperforms other retrieval baselines across different datasets due to the presence of a strong LLM backbone (Qwen2). 

\noindent Table \ref{tab:aux-res1} evaluates the retriever performance in identifying the correct source document, presenting the proportion of queries with accurate document retrieval for $k=5$. A document is considered correctly retrieved if at least $\left\lceil k/2 \right\rceil$ of the retrieved documents correspond to the ground truth source documents. We observe that ColQwen2 is better than the next closest BGE1.5 model by 4.5\%. Notably, we observe a substantial performance gap in this metric for SlideVQA, with visual models significantly outperforming text-only models. BM25 exhibits better performance than text-only models in this case, as slides typically contain sparse text, often comprising keywords that directly match between the query and context. Conversely, neural models struggle to capture semantic information effectively, as the textual content lacks complete sentences, limiting their ability to exploit contextual meaning.


\subsection{End-to-End Evaluation}
Table \ref{tab:results} presents the comparative performance of VisDoMRAG against Visual RAG, Textual RAG, and Long Context methods across multiple LLMs, including Qwen2VL (7B), Gemini Flash, and GPT-4. The results indicate that VisDoMRAG consistently achieves superior performance over the baseline methods across datasets, with performance gains ranging from 2.1-21.6\% (PaperTab), 0.67-36.14\% (FetaTab), 0.24-11.24\% (SciGraphQA), 0.81-32.87\% (SPIQA), 0.40-52.16\% (SlideVQA). Additionally, within each baseline method for most datasets, we observe a positive correlation between model size and performance, which aligns with established expectations in LLM scaling behavior \cite{Hestness2017DeepLS}.
\begin{figure}[htbp]
    \centering
    

    \begin{subfigure}[b]{0.23\textwidth}
        \centering
        \includegraphics[width=\textwidth]{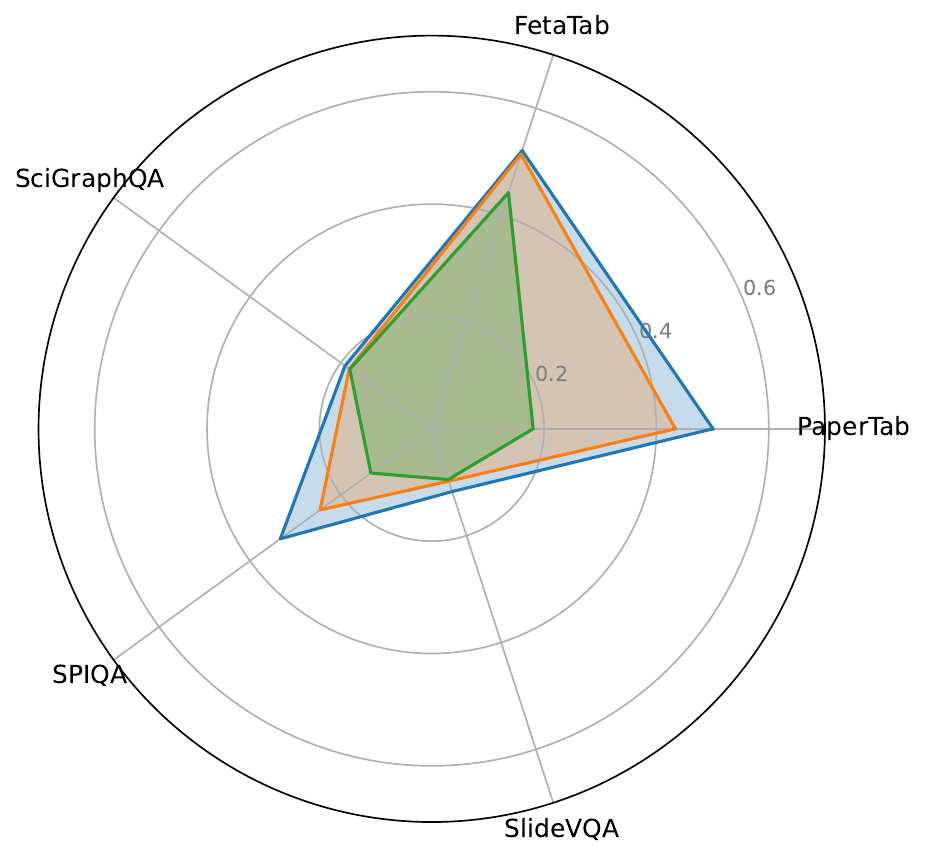}
        \caption{Long Context}
        \label{fig:visual_rag}
    \end{subfigure}
    \begin{subfigure}[b]{0.23\textwidth}
        \centering
        \includegraphics[width=\textwidth]{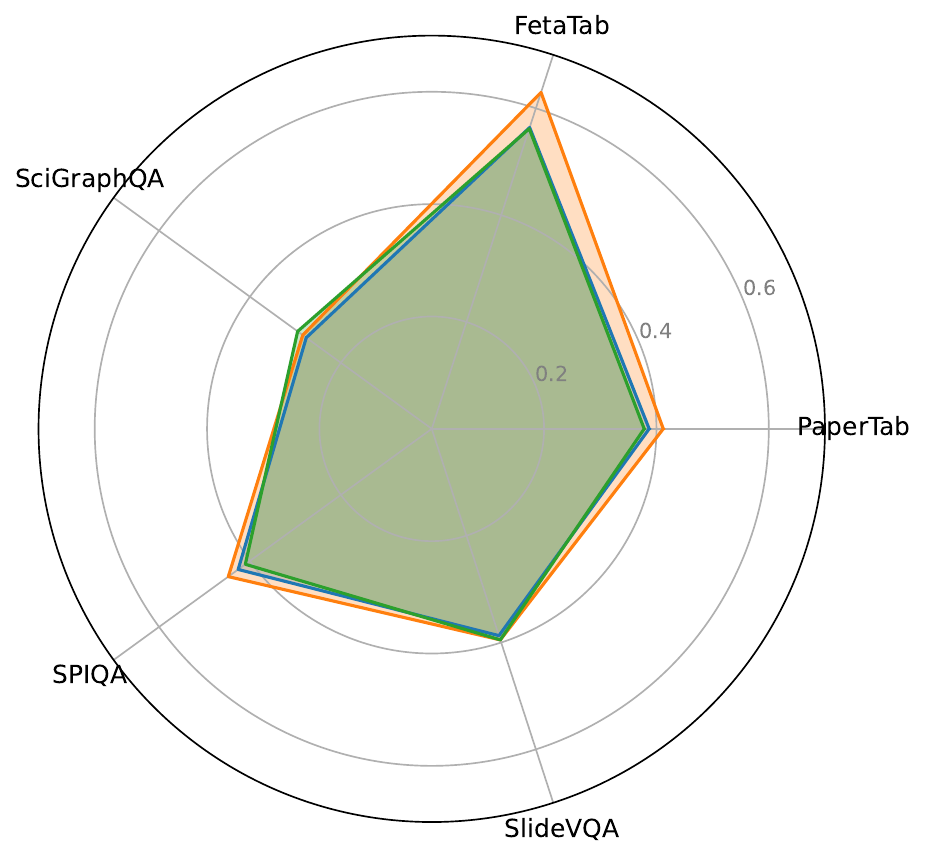}
        \caption{VisDoMRAG}
        \label{fig:visdomrag}
    \end{subfigure}
    
    \caption{Comparative performance between Long Context and \texttt{VisDoMRAG} (averaged across LLMs) evaluated on different ranges of  number of pages $\bar{p} = \sum_{d \in \mathcal{D}} |d|$, with \textcolor{cyan}{Low} ($ \bar{p} \le 100$), \textcolor{orange}{Medium} ($100 < \bar{p} \le 150$), and \textcolor{OliveGreen}{High} ($ 150 \le \bar{p}$) volumes.}
    \label{fig:radar_charts}
\end{figure}

\noindent\textbf{Textual vs Visual RAG}: In comparing the performance of textual RAG vis-à-vis visual RAG, we observe that visual RAG consistently outperforms textual RAG. This behaviour can be explained on the basis of our dataset composition which predominantly consists of visually-rich content, and visual RAG is able to leverage visual information directly. However, the performance difference is less pronounced in scientific figure datasets such as SciGraphQA and SPIQA due to the text-rich nature of scientific papers, where figures are often accompanied by detailed descriptions within the text and captions, particularly emphasizing key results and structural details. In contrast, we see a substantial performance gap between textual and visual RAG for SlideVQA, as slides typically lack extensive textual descriptions of visualizations, forcing the visual modality to be the primary source for answering questions. Additionally, we find that Gemini often performs better in the textual modality compared to the visual modality across most datasets. This disparity could be attributed to factors such as linguistic bias \cite{niu2021counterfactual, wang2024mdpo} or visual hallucination \cite{ghoshvisual}, where the model's visual perception may be less reliable than its linguistic capabilities.

\noindent\textbf{Effect of Long-Context LLMs}: We observe that \texttt{VisDoMRAG} has the ability to significantly enhance the performance of smaller models, as seen from Qwen2VL. This improvement can be attributed to its ability to integrate visual and textual reasoning, compensating for the weaker long-context understanding and visual perception. The long-context LLM baselines prove to be less effective in our setup due to the high token count and the nature of the task, which requires retrieval of specific, localized evidence—essentially a needle-in-the-haystack problem. The combination of modalities in \texttt{VisDoMRAG} mitigates these challenges, resulting in more robust answer generation, as reflected in the results.

\noindent\textbf{Effect of Increasing Page Count}: Figure \ref{fig:radar_charts} evaluate the performance of different approaches averaged across LLMs, segmented by the volume of pages associated with each query. As anticipated, long-context models exhibit significant performance drop with increasing number of pages in the collection. Contrastively, our multimodal RAG approach shows consistent QA performance even at high page counts as it is able to constrain the amount of context the LLM needs to process to answer the question effectively. 

\begin{figure}[h]
    \centering
    \includegraphics[width=\linewidth]{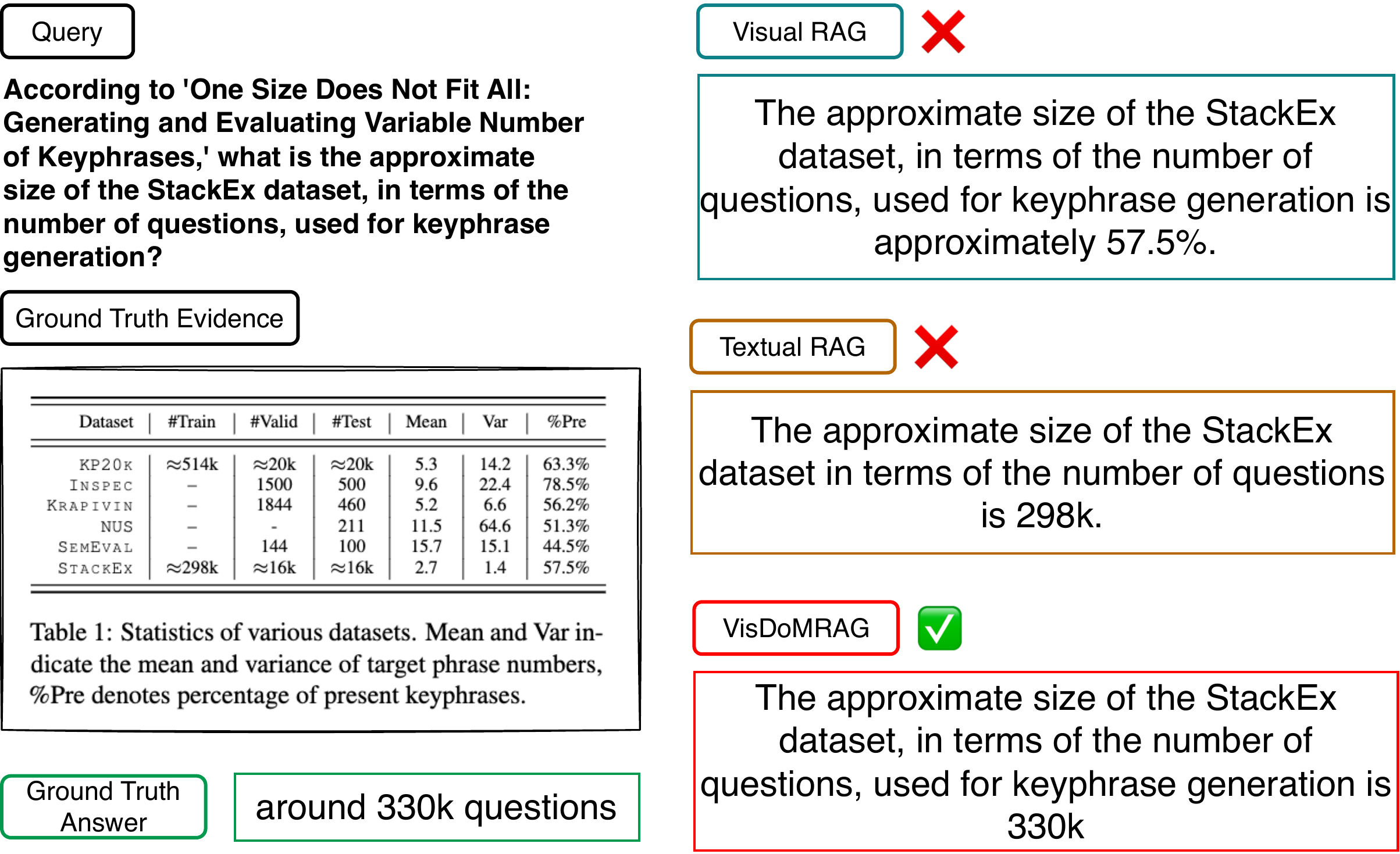}
    \caption{Qualitative example from the PaperTab dataset, comparing \texttt{VisDoMRAG} with Unimodal RAG strategies.}
    \label{fig:enter-label}
\end{figure}

\noindent\textbf{Qualitative Examples}: Fig \ref{fig:enter-label} represents a qualitative example from the PaperTab dataset, where \texttt{VisDoMRAG} effectively uses reasoning chains and answers from unimodal RAG outputs to synthesize the correct answer. More qualitative results are presented in the Appendix.

\subsection{Ablations}

We conducted ablation studies with ChatGPT4o to evaluate the effectiveness of various components in our proposed VisDoMRAG framework, as well as to compare early fusion and late fusion strategies for modality integration. The results are summarized in Table \ref{tab:results_single_column}.

\noindent\textbf{Early Fusion vs. Late Fusion:} In our experiments, early fusion, where text extracted from document pages retrieved by the visual retriever is directly appended to the visual RAG context and used as input to the LLM, demonstrated suboptimal performance compared to the late fusion strategy employed in VisDoMRAG. Specifically, early fusion struggled to integrate visual and textual evidence effectively, particularly in cross-modal reasoning, resulting in an average score of 43.63 across datasets. This limitation is likely due to the lack of independent processing for each modality, which led to weaker contextual understanding and reasoning. In contrast, late fusion—where each modality is processed independently before aggregating—proved more effective. This performance gap highlights the importance of preserving modality-specific representations before combining them, particularly when reasoning requires nuanced cross-modal evidence integration.

\noindent\textbf{Prompt Ablation:}
The ablation of our proposed prompting strategies also revealed the significance of Evidence Curation, Chain-of-Thought (CoT) prompting, and Reasoning Consistency. By replacing these components with simplified prompts that employ a basic structure where the model directly generates an answer based on the question and retrieved context, without leveraging evidence curation, chain-of-thought (CoT) prompting, or reasoning consistency mechanisms. For instance, removing these prompting strategies led to an average score drop from 37.33 to 34.68 in the text-only setting and from 49.02 to 43.93 in the vision-only setting, highlighting the importance of structured prompts.

For the VisDoMRAG setting, prompt ablation led to an average performance reduction from 50.01 to 45.98, with the most notable declines observed in datasets requiring complex reasoning, such as SPIQA and SlideVQA. The simplified prompts appeared insufficient for handling the intricacies of cross-modal evidence alignment and aggregation, leading to degraded performance in these scenarios.

\begin{table}[t]
\centering
\resizebox{\columnwidth}{!}{%
\begin{tabular}{llcccccc}
\hline
\textbf{Baseline} & \textbf{Experiment} & \textbf{PaperTab} & \textbf{FetaTab} & \textbf{SciGraphQA} & \textbf{SPIQA} & \textbf{SlideVQA} & \textbf{Average} \\ \hline
\multirow{2}{*}{{\color[HTML]{B46504} \textbf{Text}}} 
& Ours & 37.34 & 60.82 & 29.74 & 42.80 & 15.97 & 37.33 \\
& Prompt Ablation & 33.29 & 58.81 & 30.16 & 37.81 & 13.32 & 34.68 \\ \hline
\multirow{2}{*}{{\color[HTML]{0E8088} \textbf{Vision}}} 
& Ours & 42.01 & 61.89 & 31.12 & 43.28 & 66.82 & 49.02 \\
& Prompt Ablation & 34.52 & 59.85 & 31.31 & 32.55 & 61.44 & 43.93 \\ \hline
\multirow{3}{*}{{\color[HTML]{CB0000} \textbf{VisDoMRAG}}} 
& Ours & \textbf{44.11} & \textbf{63.28} & \textbf{31.36} & \textbf{44.09} & \textbf{67.22} & \textbf{50.01} \\
& Prompt Ablation & 38.34 & 62.65 & 27.85 & 36.75 & 64.33 & 45.98 \\ 
& Early Fusion & 37.37 & 61.29 & 27.94 & 33.45 & 58.12 & 43.63 \\
\hline
\end{tabular}%
}
\caption{Performance comparison of baseline approaches with ablations on \texttt{VisDoMBench}.}
\label{tab:results_single_column}
\end{table}
\section{Conclusion and Future Work}
In this work, we introduced \texttt{VisDoMBench}, the first QA dataset designed to evaluate multi-document systems incorporating visually rich elements such as tables, charts, and slides. By targeting documents that require both textual and visual comprehension, \texttt{VisDoMBench} offers a novel benchmark to assess the capability of multimodal retrieval systems. We also presented \texttt{VisDoMRAG}, a multimodal Retrieval-Augmented Generation approach that fuses visual and textual pipelines using consistency-constrained modality fusion. This method demonstrated a significant improvement over traditional long context, textual, and visual RAG by 12-20\%. While the current work focuses on RAG in multimodal multi-doc settings, future work will extend this approach to include reasoning through end-to-end trained models, especially in low-resource settings.

\section{Ethics Statement}
We use publicly available datasets in this research. The identities of human evaluators remain confidential, and no personally identifiable information (PII) is used at any stage of our experiments. Our work is solely intended for document QA applications. For a deeper understanding of potential risks and mitigation strategies in LLM safety, we direct users to relevant works by \citep{risks1, risks2, risks3}.

\section{Limitations}

Despite the advancements presented in this study, several limitations warrant consideration:

\noindent\textbf{(1) Text Extraction and Document Parsing}: A key argument for the efficacy of visual retrieval methods is the elimination of text extraction and document parsing pipelines \cite{faysse2024colpali}. However, our approach retains this overhead, which may introduce additional complexity and processing time. 

\noindent\textbf{(2) Multiple LLM calls}: Our methodology necessitates multiple LLM calls; specifically, we make three LLM calls per query. While this approach may not be optimal, it is still more cost-effective than utilizing long-context models.

\noindent\textbf{(3) Hallucinations}: As with all works involving large language models (LLMs), our approach is subject to inherent limitations related to AI safety and the risk of hallucination. These issues can affect the reliability and accuracy of the generated outputs and underscore safety risks, highlighting the need for ongoing research and refinement in the field of AI to mitigate these challenges.

\noindent Additionally, unlike previous visual QA research, which typically required models to answer questions based solely on visual data, our framework incorporates document context. This inclusion allows for relevant textual information from other sections of the paper to contribute to the query response. However, this reliance on document context represents a limitation common to all visually rich document QA datasets, as it challenges the isolation of visual performance testing. Nonetheless, this characteristic may not be entirely detrimental; in fact, it more accurately reflects the complexity of real-world systems where multimodal information is often interdependent.

\bibliography{custom}

\appendix
\section{Appendix}
\subsection{Baselines}
\subsubsection{Retrieval Models}
\paragraph{BM25} BM25 \cite{robertson1995bm25} is a widely adopted term-based ranking function based on the probabilistic information retrieval model. It calculates the relevance of a document to a given query by considering term frequency, inverse document frequency, and document length normalization. BM25 is effective for sparse text retrieval tasks, making it a standard baseline in information retrieval evaluations. We use the Python \href{https://github.com/dorianbrown/rank_bm25}{\texttt{rank\_bm25}} implementation for our experiments.

\paragraph{MiniLM} MiniLM \cite{wang2020minilmdeepselfattentiondistillation} is a lightweight, transformer-based model designed for efficient knowledge distillation. It compresses the knowledge of larger pre-trained models into a smaller architecture while maintaining competitive performance in natural language understanding tasks. MiniLM is used in retrieval tasks due to its ability to balance computational efficiency and accuracy. We use the \href{https://huggingface.co/sentence-transformers/all-MiniLM-L6-v2}{\texttt{sentence-transformers/all-MiniLM-L6-v2}} implementation in our experiments.

\paragraph{MPNet} MPNet \cite{mpnet} is a transformer-based model that leverages permuted language modeling for pre-training, which helps it capture contextual information more effectively than traditional masked language models. It excels in a variety of natural language processing tasks, including text retrieval, due to its robust contextual embeddings and representation learning capabilities. We use the \href{https://huggingface.co/sentence-transformers/all-mpnet-base-v2}{\texttt{sentence-transformers/all-mpnet-base-v2}} implementation in our experiments.

\paragraph{BGE-1.5} The BGE model family is based on a BERT-like architecture and a three-stage training process, which collectively enhance its adaptability and generalization capabilities. Pre-training is performed on large-scale plain text corpora using a tailored MAE-style approach, effectively encoding polluted text and reconstructing the clean version. The model then undergoes contrastive learning with in-batch negative sampling, leveraging large batch sizes to improve embedding discriminativeness. Finally, task-specific fine-tuning is employed using labeled datasets, applying instruction-based prompts and advanced negative sampling techniques to better accommodate diverse task types. We use the \href{https://huggingface.co/BAAI/bge-base-en-v1.5}{\texttt{BAAI/bge-base-en-v1.5}} model in our experiments, which is their large english model, version 1.5.

\paragraph{ColPali, ColQwen2} ColPali \cite{faysse2024colpali} performs late interaction retrieval on document embeddings generated directly from document page images using Vision-Language Models (VLMs). By passing the document images through PaliGemma \cite{beyer2024paligemmaversatile3bvlm}, ColPali uses the projected token embeddings to index the document pages, eliminating the need for OCR or document parsing. The multimodal alignment learned by VLMs allows both text queries and document image embeddings to exist in a shared semantic vector space, enabling more precise and efficient retrieval. ColQwen2 is a similar model with Qwen2 \cite{yang2024qwen2technicalreport} as the base VLM. We used the \href{https://huggingface.co/vidore/colpali-v1.2}{\texttt{vidore/colpali-v1.2}}, \href{https://huggingface.co/vidore/colqwen2-v0.1}{\texttt{vidore/colqwen2-v0.1}} implementations for our experiments.

\subsubsection{LLMs}
We used \href{https://huggingface.co/Qwen/Qwen2-VL-7B-Instruct}{\texttt{Qwen/Qwen2-VL-7B-Instruct}}, \href{https://platform.openai.com/docs/models}{\texttt{chatgpt-4o-latest}} and \href{https://ai.google.dev/}{\texttt{gemini-1.5-flash}} in our experiments.
For ChatGPT4o and Gemini, we set the temperature as 0.5, and use the default hyperparameters. 
For \texttt{Qwen2-VL}, the pixel range is set to $[256 \times 28 \times 28, 640 \times 28 \times 28]$. For Long Context evaluation, we use \texttt{Qwen/Qwen2-7B-Instruct} because of the implementation availability of long context inference using YaRN \cite{peng2023yarnefficientcontextwindow}. We report results on a single run of experiments.

\subsection{Datasets}

\begin{figure}[h]
    \centering
    \includegraphics[width=\linewidth]{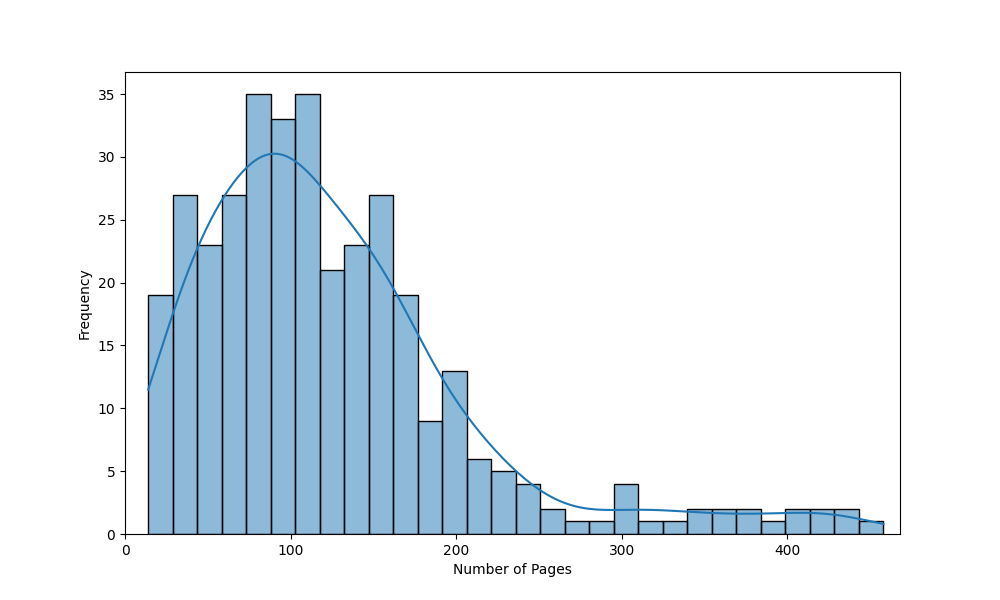}
    \caption{Distribution of pages per query for FetaTab.}
    \label{fig:dist_feta_tab}
\end{figure}

\begin{figure}[h]
    \centering
    \includegraphics[width=\linewidth]{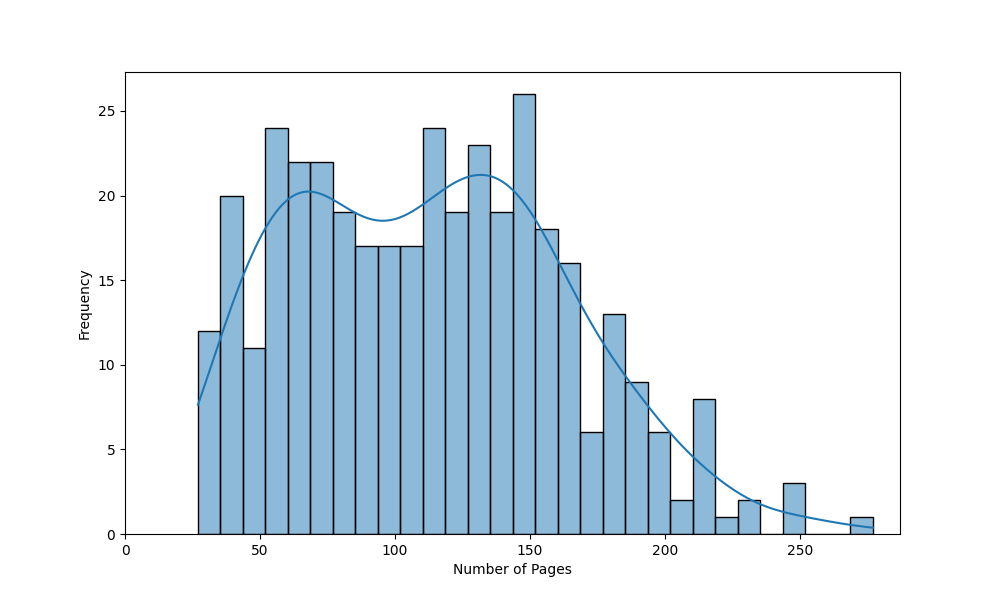}
    \caption{Distribution of pages per query for PaperTab.}
    \label{fig:dist_paper_tab}
\end{figure}

\begin{figure}[h]
    \centering
    \includegraphics[width=\linewidth]{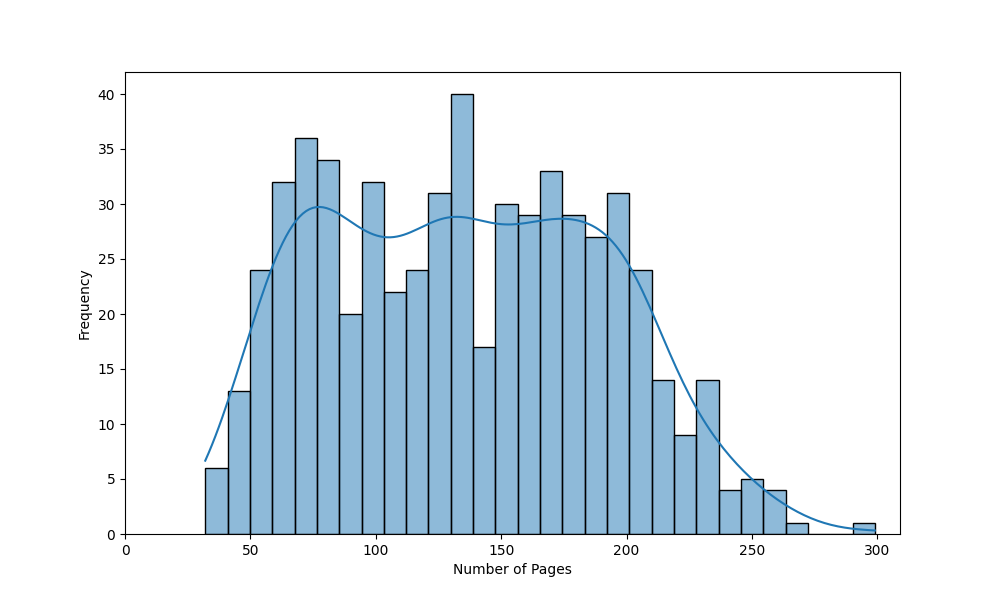}
    \caption{Distribution of pages per query for SPIQA.}
    \label{fig:dist_spiqa}
\end{figure}

\begin{figure}[h]
    \centering
    \includegraphics[width=\linewidth]{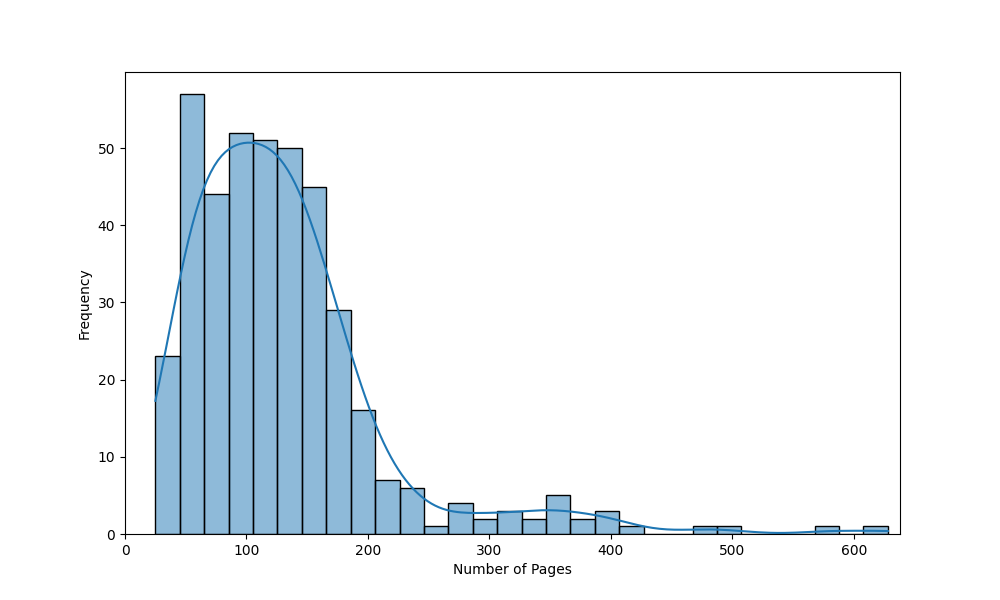}
    \caption{Distribution of pages per query for SciGraphQA.}
    \label{fig:dist_scigraphqa}
\end{figure}

\begin{figure}[h]
    \centering
    \includegraphics[width=\linewidth]{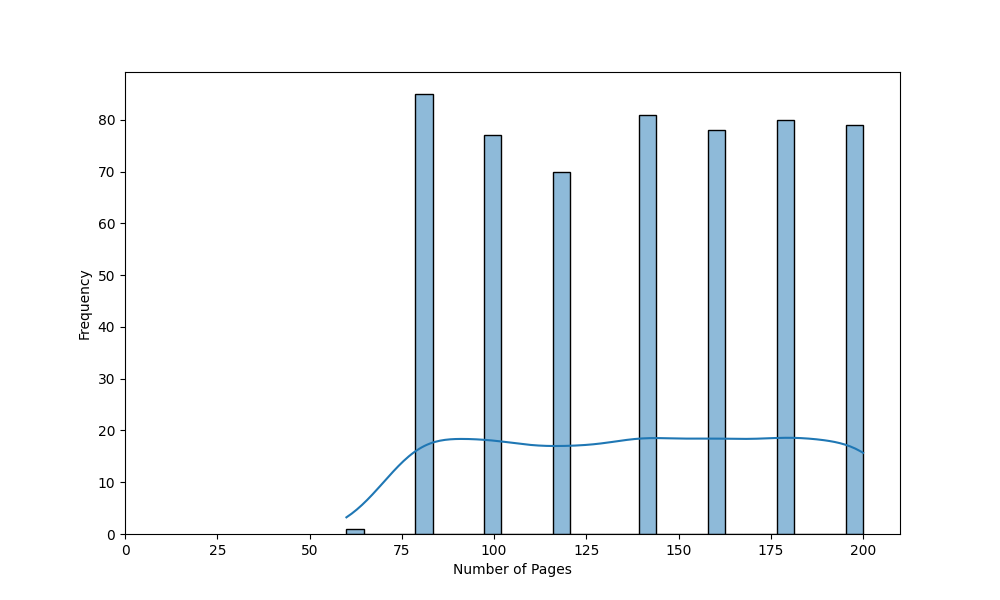}
    \caption{Distribution of pages per query for SlideVQA.}
    \label{fig:dist_slidevqa}
\end{figure}

The datasets use in our benchmark are described below.
Fig \ref{fig:dist_feta_tab}-\ref{fig:dist_slidevqa} represent the distribution of pages per query in all the data splits.

\paragraph{FetaTab} FetaTab is derived from UDA \cite{hui2024uda}, which sources its data from FetaQA \cite{nan2022fetaqa}. Many source datasets provide only segmented and partial content, lacking complete documents. To resolve this, UDA conducted a thorough source-document identification process, verifying and collecting the complete original document files based on metadata or content fragments. This was followed by rigorous matching and reorganization to form complete triplet data pairs consisting of document-question-answer. Additionally, UDA categorizes queries based on the source of factual evidence, filters out Q\&As without available answers, converts token-based data patterns to natural language, unifies data formats and structures across datasets, and designs specific LLM prompts tailored for each dataset after experimental trials. FetaTab is licensed under the CC-BY-SA-4.0 license.

\paragraph{PaperTab} PaperTab is also sourced from UDA \cite{hui2024uda}, which obtains its data from the QASPER \cite{qasper} dataset. Similar to the process described for FetaTab, UDA emphasizes the necessity of ensuring the integrity of original documents for effective document analysis. This involves a comprehensive process of identifying, verifying, and collecting complete original document files, followed by matching and reorganization to create document-question-answer triplets. UDA also categorizes queries, filters out unanswered Q\&As, converts data patterns to natural language, unifies data formats, and designs specific LLM prompts for each dataset based on experimental evaluations. PaperTab is released under the CC-BY-SA-4.0 license.

\paragraph{SPIQA} SPIQA \cite{pramanick2024spiqa} is a large-scale and challenging question-answering dataset that focuses on figures, tables, and text paragraphs extracted from scientific research papers across various computer science domains. The dataset encompasses a diverse array of visual elements, including plots, charts, schematic diagrams, and result visualizations. SPIQA consists of 270K questions divided between training, validation, and three different evaluation splits. To ensure the highest quality and reliability, SPIQA employs both automatic and manual curation methods. The dataset is released under the CC-BY-SA-4.0 license, allowing for broad use while ensuring proper attribution.

\paragraph{SciGraphQA} SciGraphQA \cite{li2023scigraphqa} is a synthetic multi-turn question-answer dataset centered on academic graphs, representing a significant advancement in the field of visual question answering. At 13 times larger than the previous largest dataset, ChartVQA, it stands as the largest open-sourced chart VQA dataset with non-synthetic charts. The dataset was constructed from 290,000 Computer Science and Machine Learning papers published on ArXiv between 2010 and 2020, with the help of Palm-2 generating 295,000 samples of open-vocabulary multi-turn question-answer dialogues about the graphs. Each dialogue is contextualized with the paper title, abstract, relevant paragraphs, and rich contextual data from the graphs, achieving an average of 2.23 question-answer turns per graph. SciGraphQA is released under the MIT license.

\paragraph{SlideVQA} SlideVQA \cite{tanaka2023slidevqa} is a multi-image document VQA dataset that contains over 2,600 slide decks, comprising more than 52,000 slide images and 14,500 questions regarding the slide content. This dataset requires complex reasoning skills, including single-hop, multi-hop, and numerical reasoning. It also provides annotated arithmetic expressions for numerical answers, enhancing numerical reasoning capabilities. More details about the dataset can be found under the license at \href{https://github.com/nttmdlab-nlp/SlideVQA?tab=License-1-ov-file#readme}{this link}.

\subsubsection{Distracting Documents}
Distracting documents are introduced as additional, irrelevant documents within the retrieval set to simulate real-world scenarios where the task is to find the most relevant context among multiple documents. These distracting documents are selected randomly from the in-domain documents of a given dataset, ensuring that they are contextually similar but not directly relevant to the query.

To validate the effectiveness of the one-to-one mapping and evaluate the robustness of the retrieval system in the presence of distracting documents, we conducted an experiment where we removed the oracle document (i.e., the ground truth document) from the retrieval set. In this setup, we provided GPT-4 with the option to refuse to answer the query if it deemed the provided context insufficient for answering the query. The refusal rate was then measured in both the default setting (with the oracle document included) and without the oracle document.

The results, shown in Table \ref{tab:refusal_rate}, reveal a significant increase in refusal rates when the oracle document is removed. In the default setting, the refusal rate is relatively low across the datasets, with PaperTab and FetaTab having 26\% and 4\% refusal rates, respectively, indicating that GPT-4 was able to find sufficient context for answering the queries. However, when the oracle document is excluded, the refusal rate jumps dramatically, with all datasets showing refusal rates between 94\% and 98\%. This increase highlights the importance of having the correct document in the retrieval set, as the model struggles to generate answers without access to the relevant context.

This experiment underscores the critical role of the oracle document in ensuring that the retrieval system can effectively answer queries and demonstrates how distracting documents can hinder retrieval performance when they introduce irrelevant or insufficient context. The results validate our approach in testing the one-to-one mapping of queries to documents and emphasize the importance of ensuring that the retrieval system can maintain performance in the presence of distracting documents.
\begin{table}[h!]
    \centering
    \resizebox{\linewidth}{!}{
    \begin{tabular}{lccccc}
        \hline
        \textbf{Method} & \textbf{PaperTab} & \textbf{FetaTab} & \textbf{SciGraphQA} & \textbf{SPIQA} & \textbf{SlideVQA} \\
        \hline
        Default & 26\% & 4\% & 18\% & 15\% & 40\% \\
        Without Oracle & 97\% & 98\% & 94\% & 97\% & 98\% \\
        \hline
    \end{tabular}}
    \caption{Refusal rate of GPT4o in the default setting and without the oracle document.}
    \label{tab:refusal_rate}
\end{table}

\subsection{Examples}

\begin{figure*}[h]
    \centering
    \includegraphics[width=\linewidth]{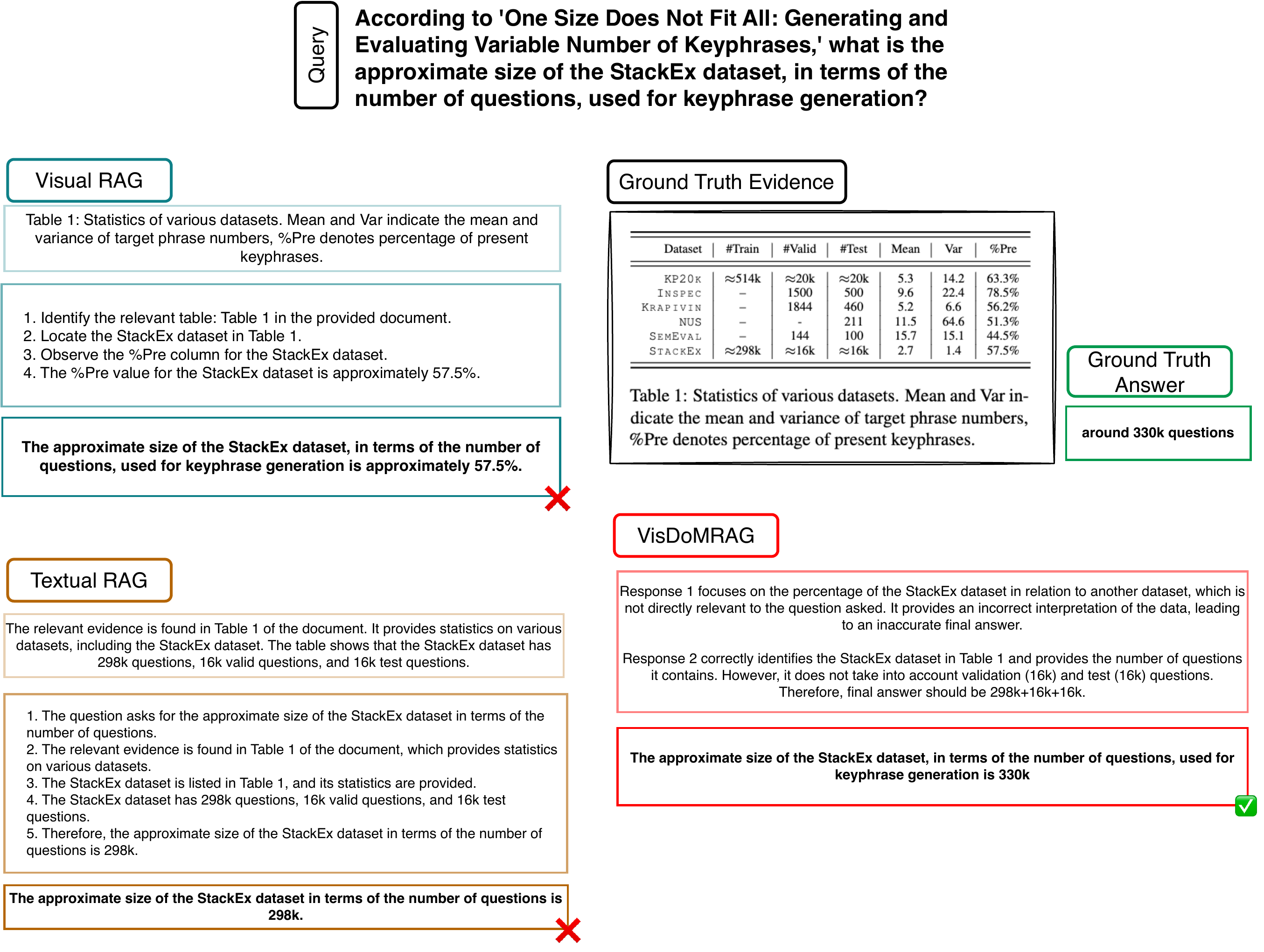}
    \caption{Qualitative example from the PaperTab dataset, comparing VisDoMRAG with unimodal RAG strategies, with Qwen2VL as the base LLM.}
    \label{fig:eg1}
\end{figure*}

\begin{figure*}[h]
    \centering
    \includegraphics[width=\linewidth]{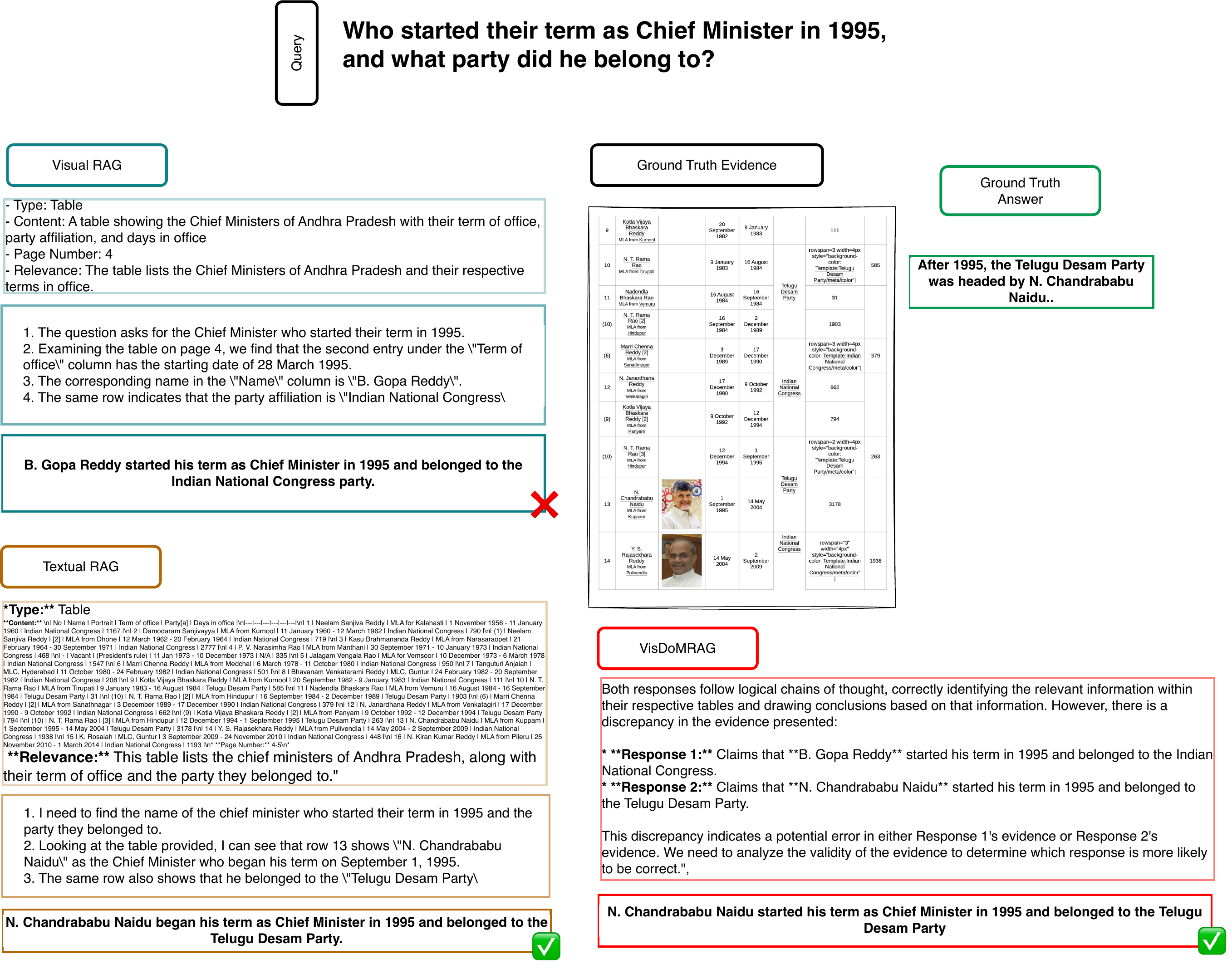}
    \caption{Qualitative example from the FetaTab dataset, comparing VisDoMRAG with unimodal RAG strategies, with Gemini as the base LLM.}
    \label{fig:eg2}
\end{figure*}

\begin{figure*}[h]
    \centering
    \includegraphics[width=\linewidth]{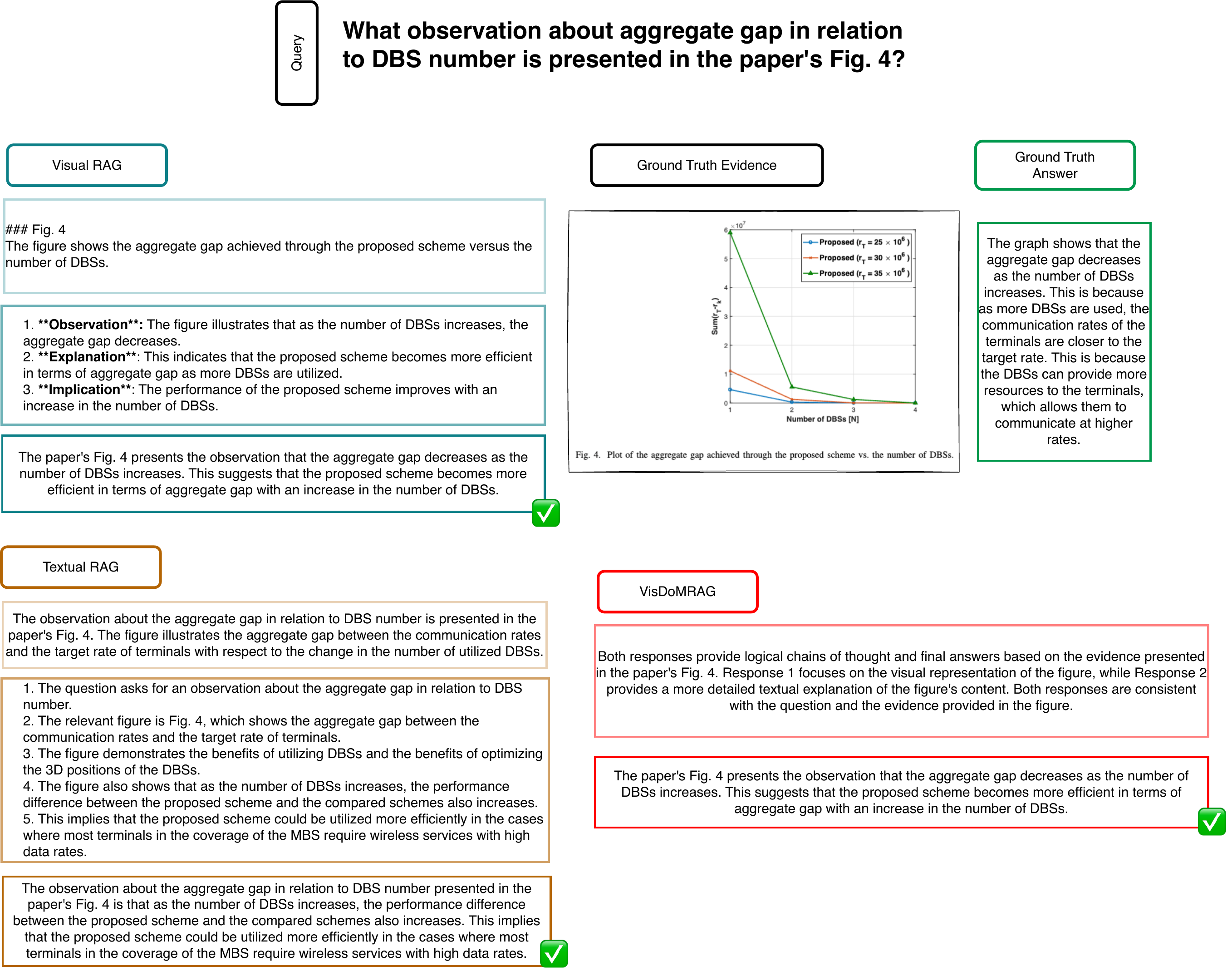}
    \caption{Qualitative example from the ScigraphQA dataset, comparing VisDoMRAG with unimodal RAG strategies, with Qwen2VL as the base LLM.}
    \label{fig:eg3}
\end{figure*}

\begin{figure*}[h]
    \centering
    \includegraphics[width=\linewidth]{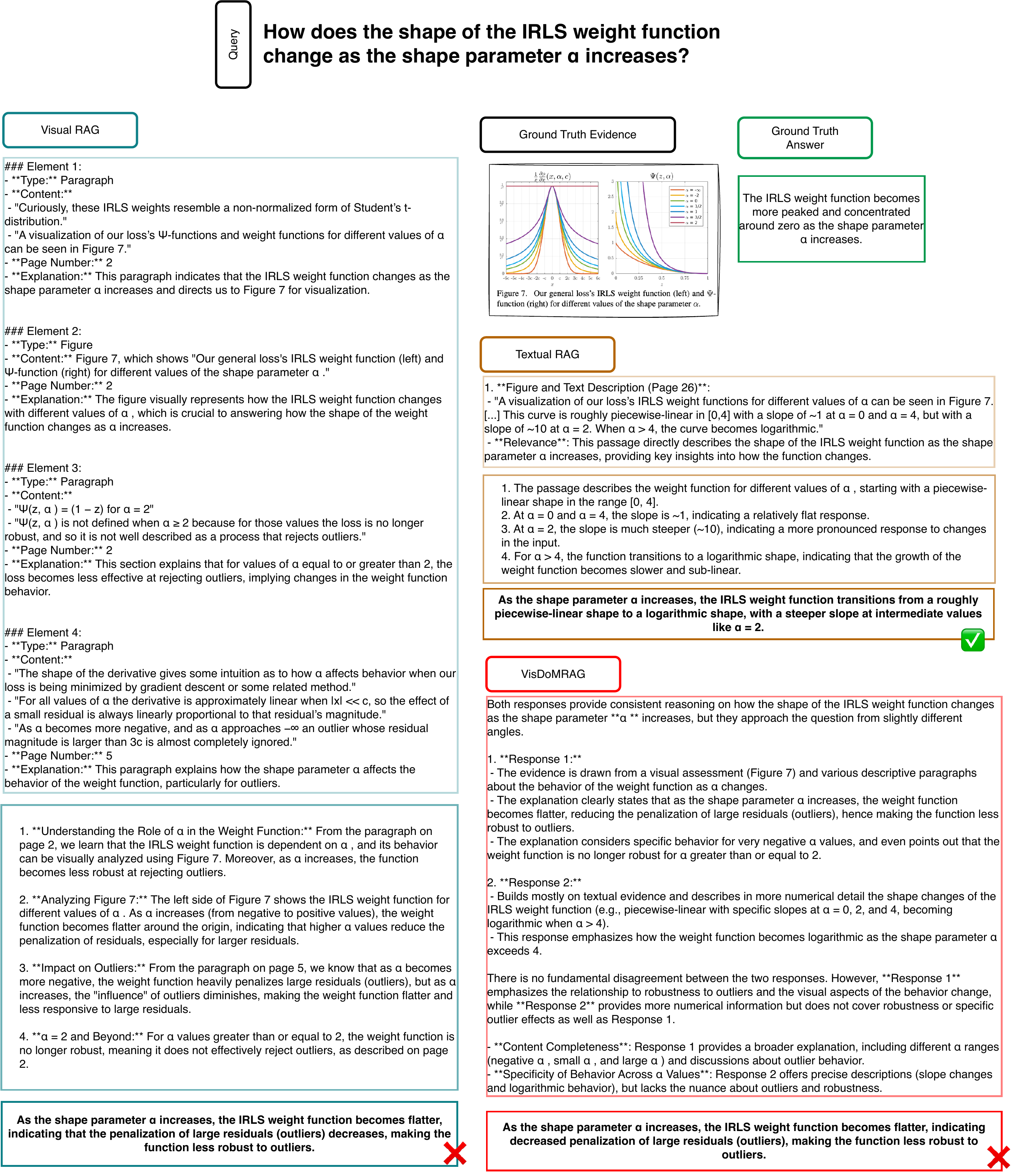}
    \caption{Qualitative example from the SPIQA dataset, comparing VisDoMRAG with unimodal RAG strategies, with ChatGPT4o as the base LLM.}
    \label{fig:eg4}
\end{figure*}

\begin{figure*}[h]
    \centering
    \includegraphics[width=\linewidth]{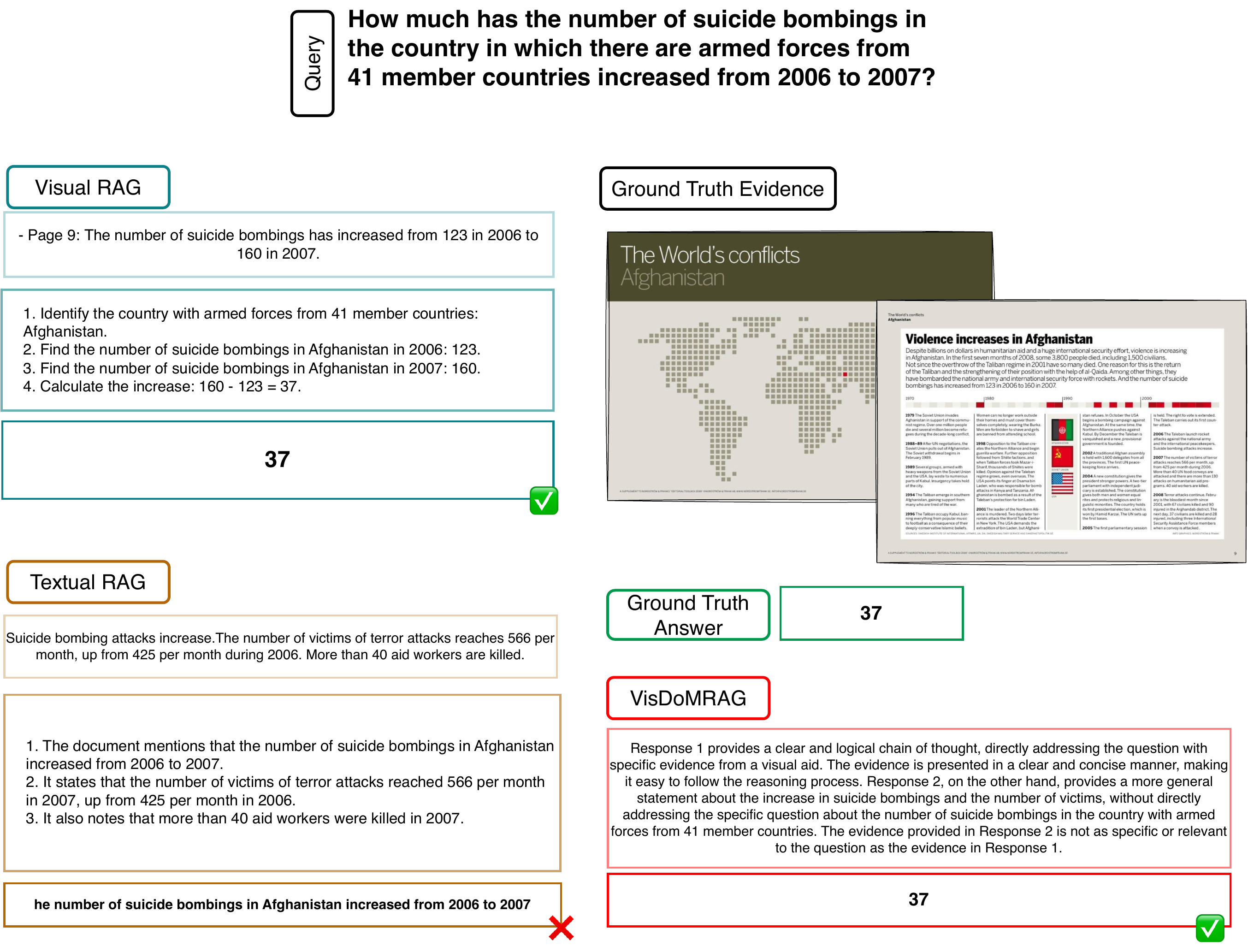}
    \caption{Qualitative example from the SlideVQA dataset, comparing VisDoMRAG with unimodal RAG strategies, with ChatGPT4o as the base LLM.}
    \label{fig:eg5}
\end{figure*}

\subsubsection{Query Augmentation}
Tables \ref{tab:papertab_queries} and \ref{tab:scigraphqa_queries} represent examples of query augmentation during dataset construction for PaperTab and SciGraphQA.
\begin{table*}[h!]
    \centering
    \resizebox{1.5\columnwidth}{!}{
    \begin{tabular}{p{0.5\textwidth} p{0.5\textwidth}}
        \hline
        \textbf{Original Query} & \textbf{Augmented Query} \\
        \hline
        What baselines did they consider? & What baseline approaches using state-of-the-art PDTB taggers were employed for the evaluation of causality prediction in the automatic causal explanation analysis pipeline? \\ \hline
        What is the average length of the claims? & What is the average token count of claims as reported in Table 2 of the PERSPECTRUM dataset? \\ \hline
        What was the performance on the self-collected corpus? & What F1 scores did the CM-Net achieve for intent detection and slot filling on the CAIS dataset as detailed in Table 6? \\ \hline
        Do they test their framework performance on commonly used language pairs, such as English-to-German? & Does the paper report results for English-to-German translation in a simulated under-resourced scenario using their proposed multilingual NMT framework? \\
        \hline
    \end{tabular}}
    \caption{Example of query augmentation from PaperTab dataset.}
    \label{tab:papertab_queries}
\end{table*}

\begin{table*}[h!]
    \centering
    \resizebox{1.5\columnwidth}{!}{
    \begin{tabular}{p{0.5\textwidth} p{0.5\textwidth}}
        \hline
        \textbf{Original Query} & \textbf{Augmented Query} \\
        \hline
        What is the main difference between the two scheduling algorithms compared in this graph? & In this paper, what scheduling algorithms are analyzed in Figure 8 for WCRT comparison? \\ \hline
        What does the phase diagram indicate about the stability of the different phases? & What does Figure 4.18 reveal about the phase boundaries for different choices of Jt and k? \\ \hline
        What does the graph show about the impact of the load-changing attack on the frequency of the system? & What does the figure show about frequency limits during the 2019 and 2020 load-changing attacks? \\ \hline
        What are some of the implications of the graph for the design of fuzzing tools? & What relationship does Fig. 3 suggest between performance and resources in fuzzing tools? \\
        \hline
    \end{tabular}}
    \caption{Example of query augmentation from SciGraphQA dataset.}
    \label{tab:scigraphqa_queries}
\end{table*}

\subsubsection{End-to-End QA Examples}

Figures \ref{fig:eg1}-\ref{fig:eg5} illustrate End-to-End QA examples across the five datasets, demonstrating the performance of different LLMs.

In Figure \ref{fig:eg1}, we analyze an example from the PaperTab dataset using Qwen2VL. VisualRAG fails in this instance by selecting the incorrect column for computation during reasoning. Conversely, TextualRAG identifies the correct column but overlooks samples from the test and validation sets. VisDoMRAG evaluates both outputs and produces the correct answer, demonstrating its ability to refine responses across modalities.

Figure \ref{fig:eg2} presents an example from the FetaTab dataset, where Gemini is employed as the base LLM. Here, TextualRAG successfully generates the correct answer by accurately verbalizing the OCR-processed table during evidence retrieval. Although VisualRAG underperforms in this case, VisDoMRAG integrates the evidence effectively, providing the overall correct answer.

In Figure \ref{fig:eg3}, an example from SciGraphQA shows both Visual and Textual RAG producing correct responses. Consequently, VisDoMRAG corroborates the correct answers, confirming the alignment between both modalities.

Figure \ref{fig:eg4} depicts a scenario from the SPIQA dataset where VisDoMRAG fails to provide the correct answer. This error arises from its bias towards the longer response generated by VisualRAG, which itself is incorrect.

Lastly, Figure \ref{fig:eg5} showcases an example from the SlideVQA dataset. In this case, TextualRAG fails to capture the necessary evidence, whereas VisualRAG successfully employs multi-hop reasoning across two slides to derive the correct answer. VisDoMRAG recognizes the precision in VisualRAG's response, favoring its consistency with the question's context.

\subsection{LLM Prompts}
\begin{figure}[h]
    \centering
    \includegraphics[width=\linewidth]{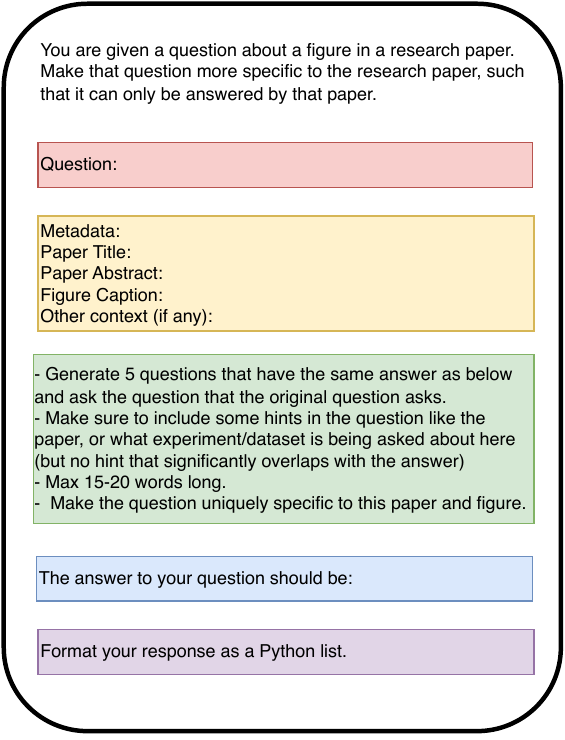}
    \caption{Prompt Template used for Query Augmentation.}
    \label{fig:prompt2}
\end{figure}

\begin{figure}[h]
    \centering
    \includegraphics[width=\linewidth]{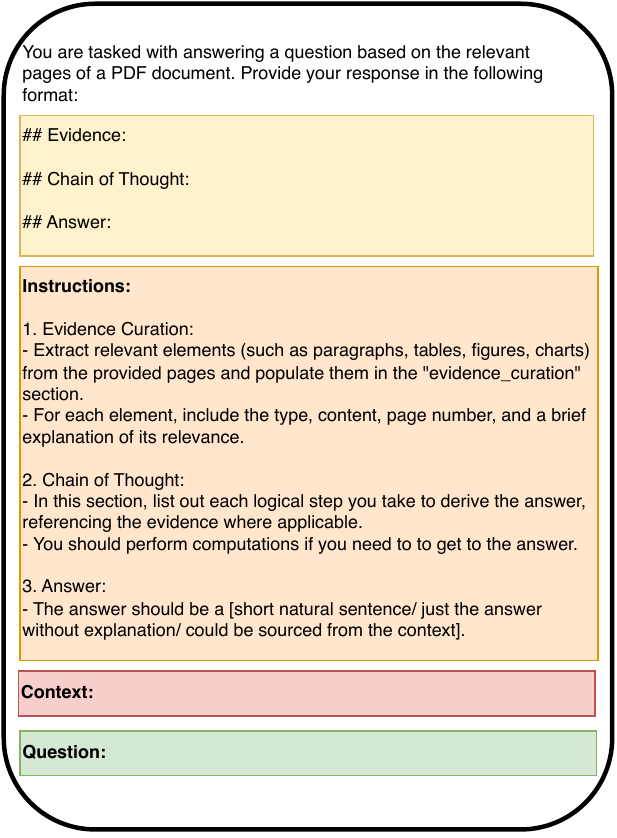}
    \caption{Prompt Template used for Unimodal RAG and Long Context experiments.}
    \label{fig:prompt1}
\end{figure}

\begin{figure}[h]
    \centering
    \includegraphics[width=\linewidth]{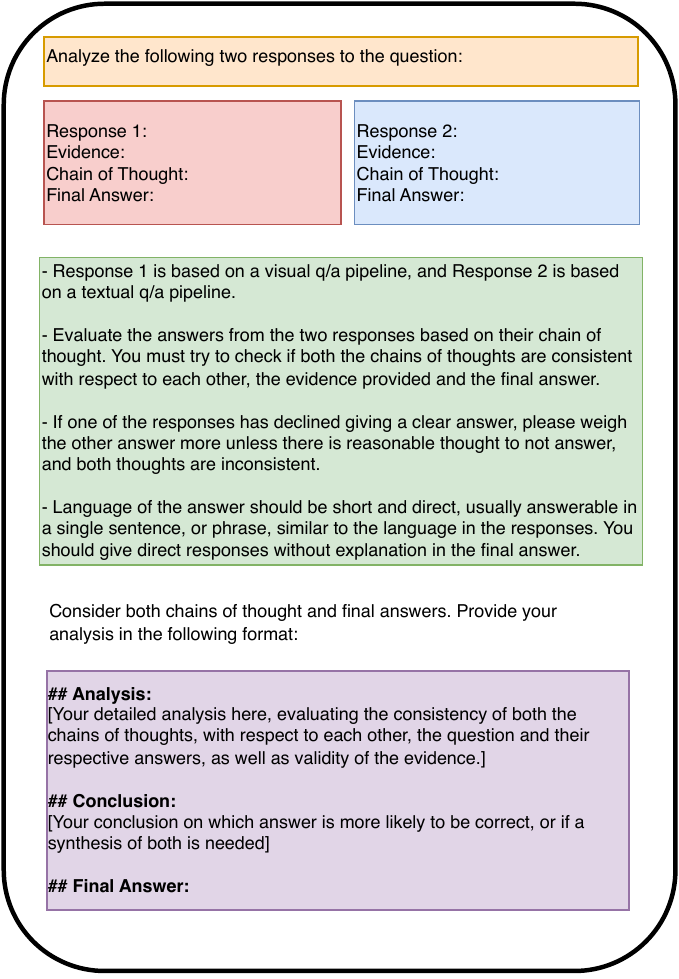}
    \caption{Prompt Template used for VisDoMRAG.}
    \label{fig:prompt3}
\end{figure}

Fig. \ref{fig:prompt2} - \ref{fig:prompt3} represent prompt templates used in our experiments for query augmentation, baselines and VisDoMRAG.

\subsection{Human Review Process}
\begin{figure*}
     \centering
     \includegraphics[width=\linewidth]{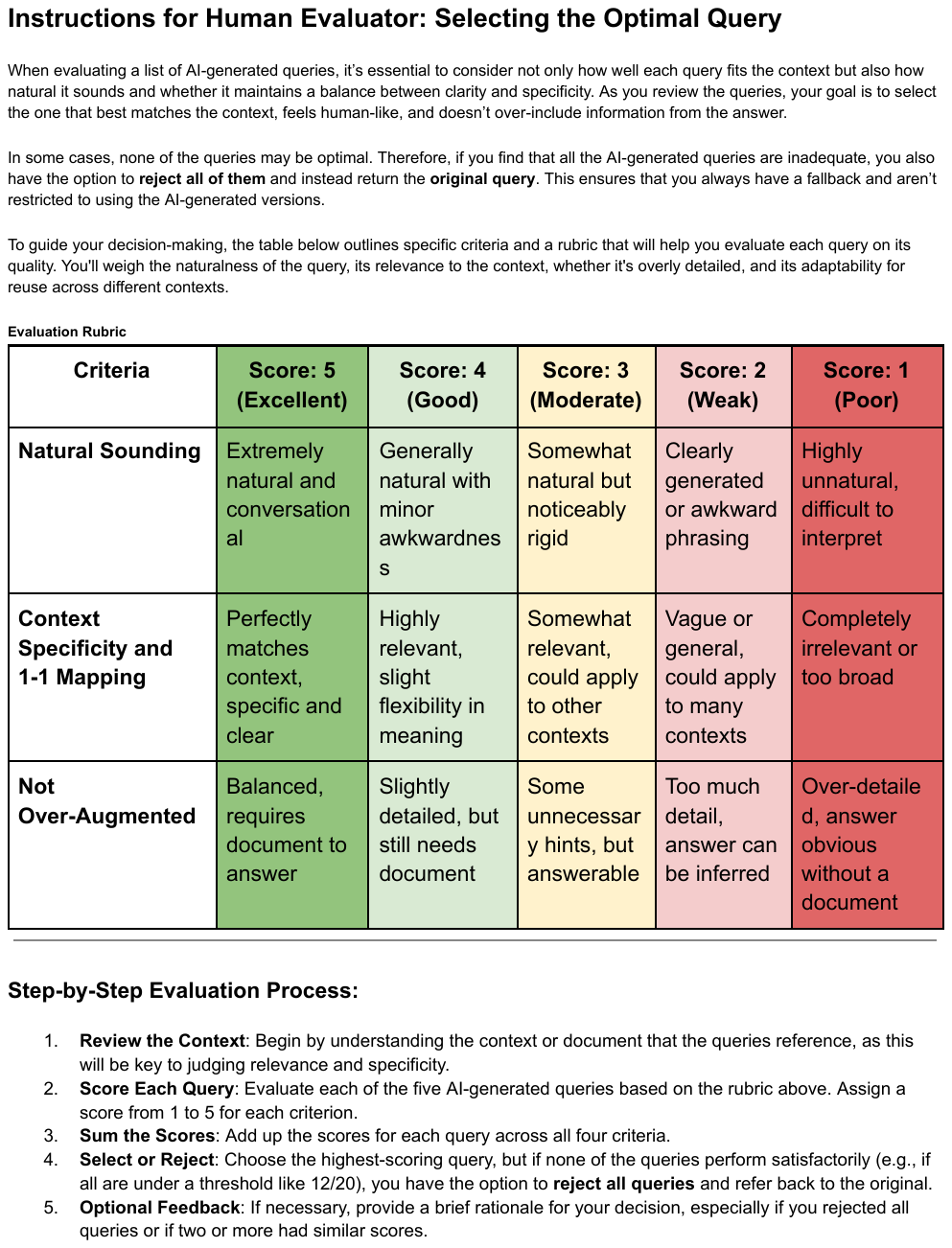}
     \caption{Brief of Reviewer Instructions, including the Evaluation Rubric.}
     \label{fig:human}
 \end{figure*}

We addressed the challenge of trivial or under-specified queries in some datasets by augmenting the queries using ChatGPT4o and relevant context, including the title and abstract of the research paper, the relevant figure's caption, and other available metadata. We employ a human reviewer to assess the quality of the generated queries and select one of the queries or reject all queries. The reviewer is a graduate student who is paid at the hourly rate for Graduate Assistants at the university where they are a student. Fig \ref{fig:human} gives a brief of the instructions as well as the evaluation rubric given to the reviewer.

\subsection{Computational Resources}
Table \ref{tab:comp} describes the Computational Resources used for running this paper's experiments. 
\begin{table}[h!]
\centering
\begin{tabular}{ll}
\hline
\textbf{Metric}     & \textbf{Details} \\ \hline
GPU Hours           & 100              \\ 
GPU Specification   & RTXA600          \\ 
Number of GPU(S) & 1 \\
Max Model Parameters      & 7B               \\ \hline
\end{tabular}
\caption{Computationa Resources for VisDoM RAG experiments.}
\label{tab:comp}
\end{table}

\end{document}